\documentclass[journal]{IEEEtran}

\usepackage{xcolor,soul,framed} 

\colorlet{shadecolor}{yellow}
\usepackage[pdftex]{graphicx}
\graphicspath{{../pdf/}{../jpeg/}}
\DeclareGraphicsExtensions{.pdf,.jpeg,.png}

\usepackage[cmex10]{amsmath}
\usepackage{array}
\usepackage{mdwmath}
\usepackage{mdwtab}
\usepackage{eqparbox}
\usepackage{url}
\usepackage{tabularx}
\usepackage{booktabs}
\usepackage{caption}
\usepackage{subfigure}
\usepackage{algorithm}
\usepackage{algorithmic}
\usepackage{hyperref}
\usepackage{svg}
\usepackage{amsmath}
\usepackage{multirow}
\usepackage{hyperref}

\begin{document}

\begin{center}
\textbf{\large IEEE COPYRIGHT NOTICE}
\end{center}

\vspace{1cm}

\begin{center}
\begin{minipage}{0.8\textwidth}
\small
\begin{quote}
\centering
\textcopyright~2023 IEEE. Personal use of this material is permitted. Permission from IEEE must be obtained for all other uses, in any current or future media, including reprinting/republishing this material for advertising or promotional purposes, creating new collective works, for resale or redistribution to servers or lists, or reuse of any copyrighted component of this work in other works.
\end{quote}
\begin{quote}
    This file corresponds to the accepted version of the manuscript in IEEE TRANSACTIONS
ON NEURAL NETWORKS AND LEARNING SYSTEMS
\end{quote}
\end{minipage}
\end{center}

\vfill


\bstctlcite{IEEEexample:BSTcontrol}
    \title{AdaptCL: Adaptive Continual Learning for Tackling Heterogeneity in Sequential Datasets}
  \author{Yuqing Zhao\IEEEauthorrefmark{1}, Divya Saxena\IEEEauthorrefmark{2}, \textit{Member, IEEE}, and Jiannong Cao\IEEEauthorrefmark{3}, \textit{Fellow, IEEE}\\
  \IEEEauthorrefmark{1}\href{csyzhao1@comp.polyu.edu.hk}{csyzhao1@comp.polyu.edu.hk}  
  \IEEEauthorrefmark{2}\href{divsaxen@comp.polyu.edu.hk}{divsaxen@comp.polyu.edu.hk}
\IEEEauthorrefmark{3}\href{csjcao@comp.polyu.edu.hk}{csjcao@comp.polyu.edu.hk}
  \\
Department of Computing, The Hong Kong Polytechnic University

}


\maketitle

\begin{abstract}
Managing heterogeneous datasets that vary in complexity, size, and similarity in continual learning presents a significant challenge. Task-agnostic continual learning is necessary to address this challenge, as datasets with varying similarity pose difficulties in distinguishing task boundaries. Conventional task-agnostic continual learning practices typically rely on rehearsal or regularization techniques. However, rehearsal methods may struggle with varying dataset sizes and regulating the importance of old and new data due to rigid buffer sizes. Meanwhile, regularization methods apply generic constraints to promote generalization but can hinder performance when dealing with dissimilar datasets lacking shared features, necessitating a more adaptive approach. In this paper, we propose AdaptCL, a novel adaptive continual learning method to tackle heterogeneity in sequential datasets. AdaptCL employs fine-grained data-driven pruning to adapt to variations in data complexity and dataset size. It also utilizes task-agnostic parameter isolation to mitigate the impact of varying degrees of catastrophic forgetting caused by differences in data similarity. Through a two-pronged case study approach, we evaluate AdaptCL on both datasets of MNIST Variants and DomainNet, as well as datasets from different domains. The latter include both large-scale, diverse binary-class datasets and few-shot, multi-class datasets. Across all these scenarios, AdaptCL consistently exhibits robust performance, demonstrating its flexibility and general applicability in handling heterogeneous datasets.
\end{abstract}


\IEEEpeerreviewmaketitle

\section{Introduction}

\begin{figure}
  \begin{center}

  \subfigure[{Traditional Parameter Isolation based Methods
}]{\includegraphics[width=\linewidth]{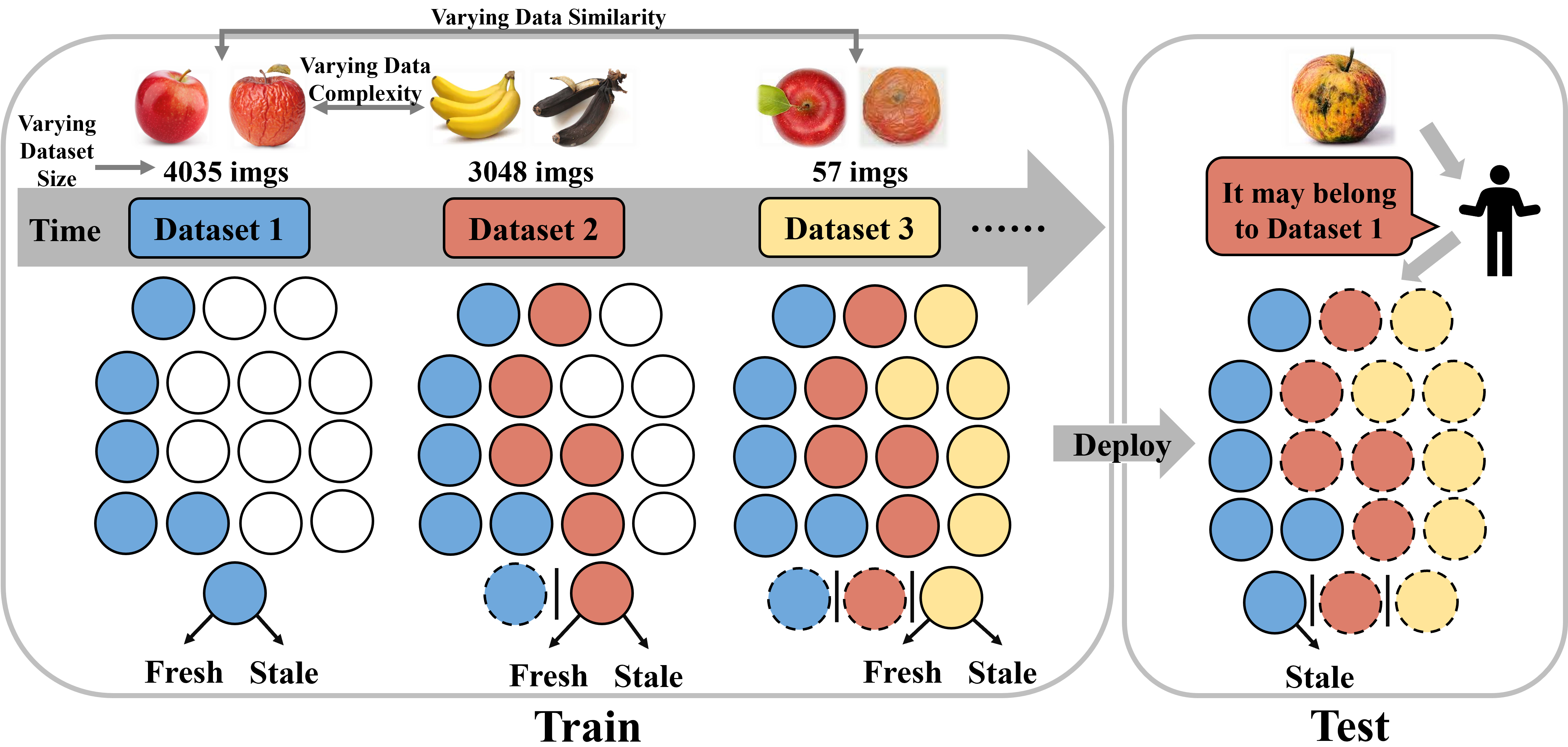}}
  \subfigure[{AdaptCL: Adaptive Learning with Task-Agnostic Parameter Isolation
}]{\includegraphics[width=\linewidth]{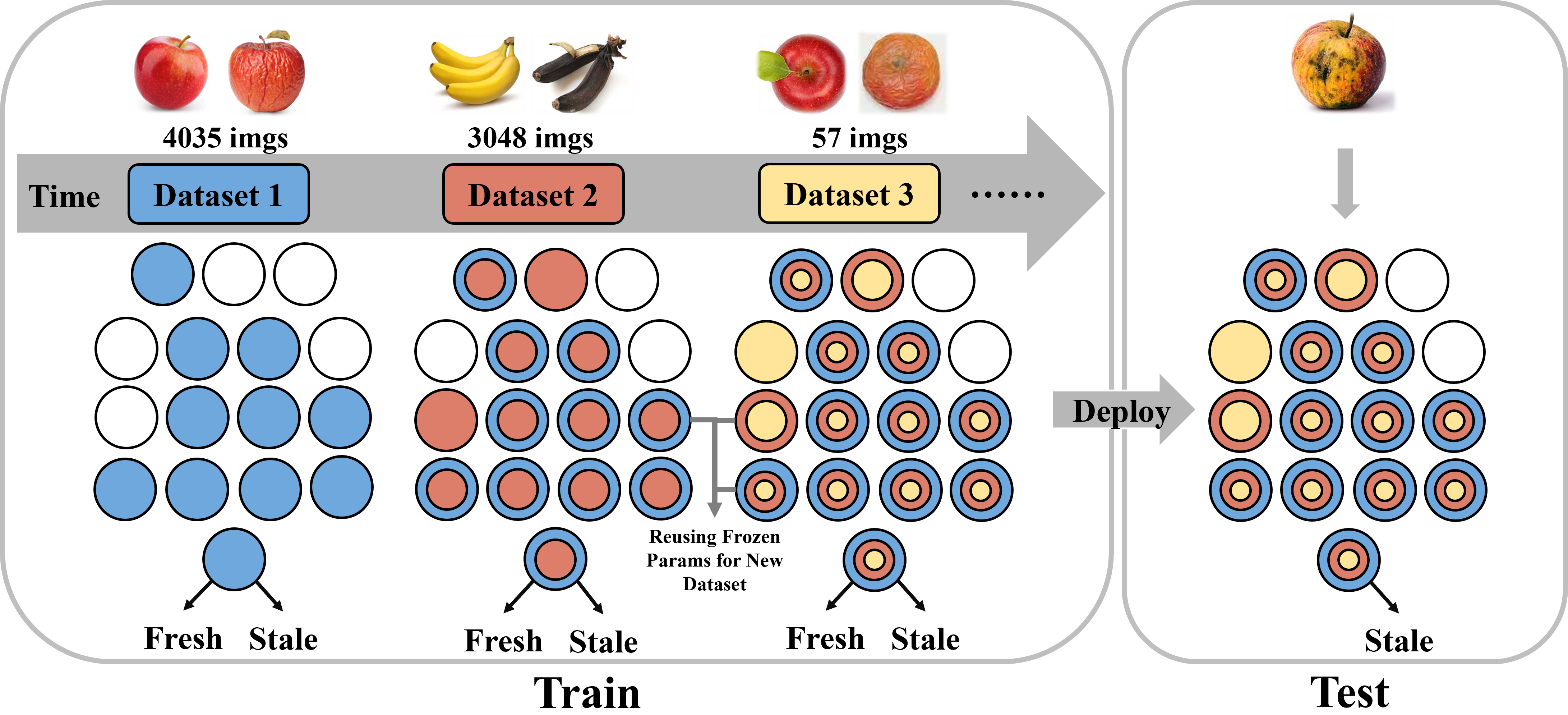}}\\
  \caption{(a) Traditional parameter isolation methods divide the network into non-interfering modules during inference. However, these methods are limited to task-specific continual learning (aka task incremental learning). They require manual selection of output layers and parameters, resulting in limited generalization and higher parameter usage.
(b) AdaptCL achieves task-agnostic parameter isolation by fine-grained data-driven parameter partitioning, enabling high accuracy on heterogeneous datasets without module selection, while also optimizing parameter reuse and saving resources.
}
     \label{figure1}
  \end{center}
\end{figure}

\IEEEPARstart{T}{he} past decade has witnessed a surge in data generation, facilitated by sensor-equipped devices and rapid digitization, across diverse domains such as healthcare, smart manufacturing, transportation, food safety, etc. However, datasets associated with these domains often originate from multiple sources or at different times, contributing to their inherent heterogeneity, which encompasses variations in dataset size, complexity, and similarity. This heterogeneity presents unique challenges, particularly in implementing continual learning algorithms.

As machine learning models, particularly continual learning models, gain prominence in these domains, it becomes evident that they must be robust and flexible enough to accommodate the inherent heterogeneity of datasets. This heterogeneity often manifests in several ways: the size of the dataset can range from few-shot examples to large-scale samples; the complexity of data can differ based on the range and intricacy of features; and the similarity of data can vary, which can create difficulties in distinguishing task boundaries. Conventional continual learning methods for these scenarios \cite{he2021online,zhao2021memory,he2022exemplar} are typically task-agnostic and depend on either rehearsal or regularization techniques, and they have limitations when dealing with such datasets. The rehearsals often struggle with size variability due to a rigid buffer size that makes the importance regulation between old and new data challenging, while regularization techniques may hinder performance when dealing with dissimilar datasets that lack shared features. These challenges underscore the need for a more adaptive approach to handling heterogeneous datasets.

On the other hand, structure-based methods like parameter isolation show great promise in handling both similar and dissimilar domains. These methods segment the network into distinct modules that do not interfere with each other during inference \ref{figure1}. However, they are primarily suitable for task-specific continual learning, where the manual selection of the parameter module, based on the task category during inference, is feasible. In the case of task-agnostic continual learning, using all parameters for integrated inference can lead to significant interference and a drop in accuracy \cite{abati2020conditional}. Therefore, the direct application of parameter isolation to task-agnostic continual learning is unsuitable without an appropriate adaptation mechanism.

Building on our previous work \cite{zhao2022memory}, we propose Adaptive Continual Learning (AdaptCL). AdaptCL enables adaptive learning through fine-grained data-driven pruning, effectively responding to variations in data complexity and dataset size. It also employs task-agnostic parameter isolation to ensure optimal model performance across datasets with varying similarity levels, all without the need for manual module selection. AdaptCL draws inspiration from the human brain's adaptive nervous system, a complex neural network that dynamically prunes redundant synapses \cite{pascual2005plastic,johnston2009plasticity} and reuses neural circuits for different tasks without compromising the original functions during development \cite{anderson2010neural}.

AdaptCL is a pioneering task-agnostic continual learning approach designed specifically to tackle heterogeneity in sequential datasets. The key contributions of this research can be summarised as follows:

\begin{itemize}
\item We conduct the first comprehensive investigation into adaptive continual learning for managing heterogeneous datasets, irrespective of their complexity, size, and similarity. This innovative approach does not require retraining or different models for varying batches of data, marking a significant leap in continual learning techniques.

\item Our method, AdaptCL, uniquely employs a combination of fine-grained data-driven pruning and task-agnostic parameter isolation to address the problems of catastrophic forgetting and variations in data complexity and dataset size. These adaptive mechanisms enable the model to respond effectively to different scenarios, increasing both flexibility and robustness.

\item Extensive experiments conducted on several datasets, including MNIST Variants, DomainNet, and large-scale diverse and few-shot, multi-class food quality datasets, demonstrate the general applicability and resilience of AdaptCL. The method consistently outperformed existing solutions, providing higher average accuracy and versatility across different networks and applications.

\end{itemize}


\section{Related Works}
Continual learning methods are crucial tools in the field of machine learning, aiding in the effective handling of tasks that evolve over time. The existing methods primarily fall into three categories: rehearsal-based, regularization-based, and structure-based. This section provides a detailed overview of these methods, highlighting their strengths and limitations, particularly when applied to task-agnostic continual learning and the management of heterogeneous datasets.

\subsection{Task-Agnostic Continual Learning}

\subsubsection{Rehearsal-Based}
These techniques seek to overcome catastrophic forgetting, a significant challenge in continual learning, by replaying previous training data periodically. Early methods like GEM and A-GEM \cite{lopez2017gradient, chaudhry2018efficient} relied on storing a portion of past training data and reusing it in future training phases.
This approach has been further refined by Peng et al. \cite{10197260} and Ho et al. \cite{10058177} with the incorporation of generative networks to create synthetic data distributions for pseudo-rehearsals. To enhance memory efficiency in rehearsal techniques, Zhao et al. \cite{zhao2021memory} adopted auxiliary low-fidelity exemplar samples.

LwF \cite{li2017learning} introduces knowledge distillation that utilizes a teacher network to distil knowledge and soft targets to a student network while training on new tasks, enabling retention of knowledge from previous tasks.
Some combine replay with knowledge distillation like Andrea et al. \cite{rosasco2022distilled} keep a very small buffer for highly informative samples and combine with distillation playback and Jingyuan et al. \cite{sun2020distill} distils knowledge and replays experience from previous tasks when fitting on a new task.
ICaRL \cite{rebuffi2017icarl} adopts a combination of rehearsal and regularization through learning a compact and discriminative feature representation to enable class-incremental learning.
Similarly, \cite{he2022exemplar} adopts a combination of rehearsal and regularization that uses the nearest class mean (NCM) classifier on food image classification dataset Food1k-100; the class mean of all data seen so far is estimated by the online mean update standard during the training phase.
PRE-DFKD\cite{binici2022robust} further refines these strategies and proposes to rehearse the model using the data-free knowledge distillation through the distribution of the previously observed synthetic samples from a Variational Autoencoder.

Despite these advancements, rehearsing techniques face limitations when managing datasets of varying sizes and maintaining the balance between old and new data. However, with AdaptCL, the model allocates parameters based on the accuracy in a data-driven way, allowing it to retain knowledge as parameter-level representations, independent of the data volume.

\subsubsection{Regularization-Based}
These methods incorporate regularization techniques, such as weight decay or dropout, to prevent catastrophic forgetting in neural networks when learning multiple tasks sequentially.
Inspired by Bayesian Learning, Elastic Weight Consolidation (EWC) \cite{kirkpatrick2017overcoming,huszar2017quadratic} mitigates catastrophic forgetting by tracking changes using the Fisher Information Matrix.

\cite{he2021online} adopts knowledge distillation on augmented exemplars in a class-incremental setting on food image classification.

Liu et al. \cite{9705128} introduces an instance neighborhood-preserving loss and a label priority-preserving loss to maintain the relative relationships of model responses within a set of instances and the relative relationships of model outputs with respect to an instance, contrasting with traditional knowledge distillation methods that retain absolute responses of isolated instances.

P\&C \cite{schwarz2018progress} compress learned knowledge and distil it into the knowledge base, and preserve knowledge with EWC while using the active column to progress new data. Using a Bayesian neural network, CBLN \cite{li2020continual} preserves distinctive parameters for different datasets for retaining performance. Similarly, \cite{park2020convolutional} introduced developmental memory (DM) into a CNN, continually growing sub-memory networks to preserve important features of learned tasks while allowing faster learning. Each sub-memory can store task-specific knowledge by using a memory loss function and preserve it during continual adaptations.
HAT \cite{serra2018overcoming} learns an attention mask over important parameters. By aligning local representations, P-TNCN \cite{ororbia2020continual} replaces the back-propagation method that descent steepest, punishing parameter updates to a more generalized result, therefore mitigating catastrophic forgetting.

Despite the potential of regularization-based methods, they can face challenges when handling heterogeneous datasets, especially those that are dissimilar and have few shared features. While through parameter isolation in a data-driven manner, AdaptCL can effectively adapt to datasets with varying levels of similarity, including dissimilar ones.

\subsection{Task-Specific Continual Learning}

\subsubsection{Structure-Based}

Structure-based methods are primarily employed in task-specific scenarios, and these methods use parameter isolation to handle both similar and dissimilar domains effectively. They divide the network into separate modules to mitigate interference during inference. While these techniques excel in managing catastrophic forgetting, they present difficulties when directly applied to task-agnostic scenarios.

One approach, exemplified by Progressive Neural Nets (PNNs) \cite{rusu2016progressive}, involves a static growth of the architecture with equal-sized modules, allowing for forward knowledge transfer between them. However, this method lacks a data-driven approach and requires task-specific settings for subsequent tasks, limiting its flexibility.
Another approach, represented by SILF \cite{10036133}, addresses parameter isolation by pruning unimportant parameters, isolating the important ones to mitigate forgetting. However, SILF relies on manual pruning ratio setting instead of leveraging a data-driven approach.
Reinforced Continual Learning (RCL) \cite{xu2018reinforced} expands each layer using reinforcement learning and enables parameter sharing. Nevertheless, this method necessitates task labels as additional inputs during inference to determine the parameters to use.
To strike a balance between knowledge transfer and catastrophic forgetting, CLAW \cite{adel2019continual} identifies which parts of the network should be shared or preserved for specific tasks.
PathNet \cite{fernando2017pathnet} and RPS-Net \cite{rajasegaran2019random} adopt a modularized network with multiple possible paths from input to output. They choose specific paths based on tasks or dataset labels. Additionally, RPS-Net includes a distillation loss and retrospection replay to further minimize forgetting.
CAT \cite{ke2020continual} masks used parameters and blocks gradient flow through unused units for dissimilar tasks. Task masks are stored according to task ID or label and need to be retained during testing.
Other methods, such as DAM \cite{rosenfeld2018incremental}, CLNP \cite{golkar2019continual}, and PackNet \cite{mallya2018packnet}, leverage pruning to strike a balance between model sparsity and performance. DAM assigns learning of each domain to a fraction of the network, typically with the same percentage (e.g., 13\%). CLNP and PackNet prune parameters based on specific percentages.

Notably, the power of structure-based parameter isolation methods like PackNet has been demonstrated through recent advancements \cite{delange2021continual} that have shown superior performance compared to other continual learning methods \cite{rebuffi2017icarl, lopez2017gradient, serra2018overcoming, aljundi2018memory, kirkpatrick2017overcoming, li2017learning}. 

However, challenges persist, particularly when dealing with heterogeneous datasets in task-agnostic settings, which calls for more adaptive approaches. 
Our previous work \cite{zhao2022memory} priorly applied the structure-based parameter isolation method to the task-agnostic scenario. However, its coarse-grained pruning resulted in limited adaptability to the heterogeneity of dataset size and similarity, leading to sub-optimal accuracy. Additionally, more adequate validation is needed on heterogeneous datasets.

\section{Problem setting and objective}

We are given a sequence of non-IID datasets ${D_1,D_2,...D_n }$ for a fixed task. Each dataset consists of a group of labelled data $(X,Y)\in D$, where $X$ and $Y$ are input variables and the corresponding output variables, respectively.
A task-agnostic continual learning setting aims to optimize:
\begin{equation} \label{eq0}
\max_\theta E_{t\sim D} [E_{(X,Y)\sim D_t} [\log p_\theta (Y|X)]]
\end{equation}
where $\theta$ identifies the parametrization of the network.
Such a maximization problem is subject to continual learning constraints: when accessing the current dataset $D$ at time $t$, it is impractical or impossible to access any previous or future dataset. We aim to develop a task-agnostic continual learning method that can effectively handle a sequence of heterogeneous datasets.

Here, the absence of known task or dataset labels prevents task-aware inference in the model. The task-agnostic setting requires merging the output units into a single-headed classifier, with more serious task interference between data from different domains, which leads to more severe forgetting \cite{abati2020conditional}.

\section{Adaptive Continual Learning}

\begin{figure*}[ht]
 \centering
 \includegraphics[width=5in]{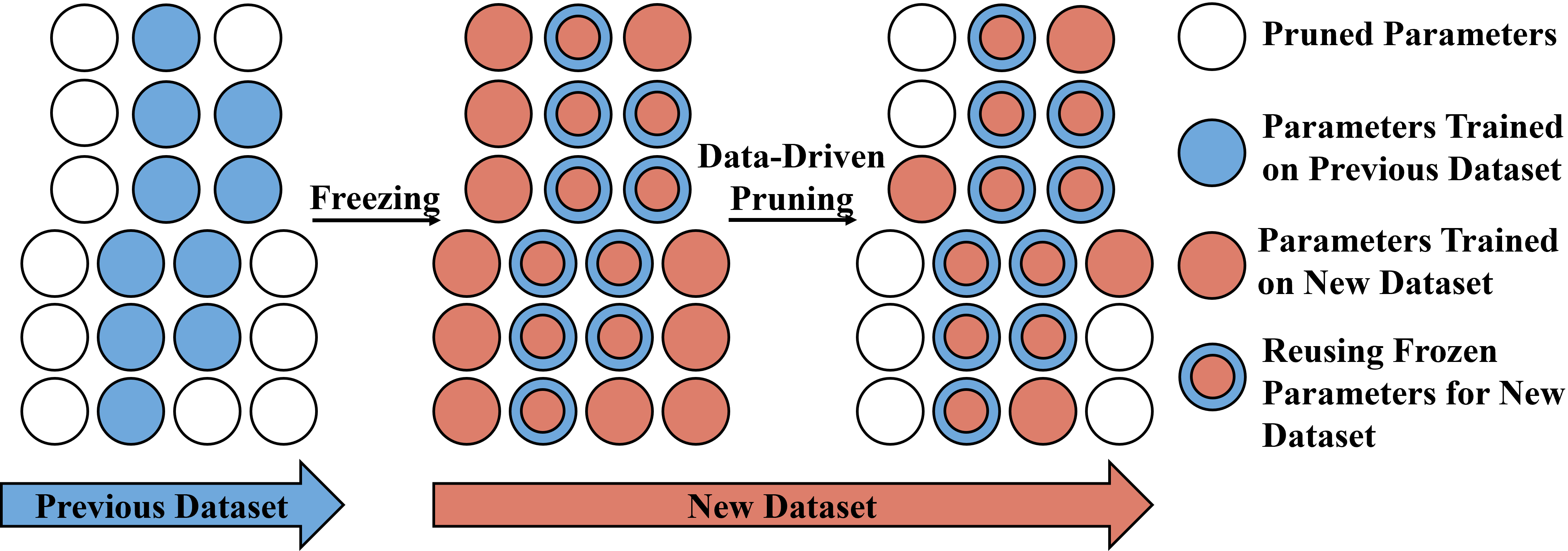}
  \caption{The Adaptive Continual Learning (AdaptCL) training flow. It facilitates adaptive learning via fine-grained data-driven pruning to respond effectively to variations in data complexity and dataset size. Additionally, it enables task-agnostic parameter isolation to ensure optimal model performance on datasets ranging in similarity without requiring manual selection of modules.}

 \label{figure2}
\end{figure*}

AdaptCL (Figure \ref{figure2}) employs adaptive learning that utilizes fine-grained data-driven pruning to adapt to variations in data complexity and dataset size. 
It also employs a form of task-agnostic parameter isolation to mitigate the impact of varying degrees of catastrophic forgetting caused by differences in data similarity.

\subsection{Fine-Grained Data-Driven Pruning}

In continuous learning with heterogeneous datasets, effectively managing model complexity becomes crucial within a limited computational budget. Fine-grained pruning goes beyond traditional pruning approaches by compressing the model while maintaining or even increasing accuracy. This data-driven pruning method aims to strike a balance between network accuracy and sparsity, facilitating better parameter reuse among similar datasets and improved fitting accuracy for complex or dissimilar datasets.
Let's consider a neural network with a parameter set $\{W_i:1\le i\le C\}$, where $W_i$ represents the parameter matrix at layer $i$ and $C$ denotes the number of layers. For fully connected layers, the corresponding parameter is $W_i\in R^{c_o\times c_i}$, where $c_o$ is the output dimension and $c_i$ is the input dimension. For convolutional layers, a convolution kernel $K_i\in R^{c_o\times c_i\times w\times h}$ exists, where $c_o$ represents the number of output channels and $c_i$, $w$, and $h$ denote the number of input channels, width, and height respectively.
Pruning involves applying a binary mask $M^p$ to each parameter $W$, setting unimportant parameters to 0. To determine the masks, a trainable pruning threshold vector $t$ is introduced. The magnitude of parameters is compared to the corresponding threshold values using a unit step function $S(x)$, as shown in equation \ref{eq2}.

\begin{equation} \label{eq1}
S\left(x\right)=    \left\{
\begin{array}{ll}
      0, & x < 0 \\
      1, & x\geq 0 \\
\end{array} 
\right. 
\end{equation}

\begin{equation} \label{eq2}
{M^p}_{ij}=S\left(\left|W_{ij}\right|-t_i\ \right),\ \ 1\le i\le c_o,1\le j\le c_i
\end{equation}
The corresponding element in pruning mask ${M^p}_{ij}$ will be set to 0 if $W_{ij}$ needs to be pruned.


Unlike traditional methods that use a fixed threshold value, achieving fine-grained pruning requires a high-dimensional threshold, denoted as $t$, in order to ensure more precise pruning. 
For a fully connected layer or recurrent layer with a parameter size of $W \in R^{c_o\times {c_i}}$, our threshold tensor size is $t \in R^{c_o}$.
Each weight $W_{ij}$ will have a neuron-wise threshold, denoted as $t_i$, where $W_{ij}$ represents the $j$th weight associated with the $i$th output neuron. Similarly, for convolutional layers, the thresholds are filter-wise.
Consequently, each neural network layer will be pruned based on high-dimensional thresholds, where each row of the tensor has its unique threshold. This approach ensures a more fine-grained pruning, avoiding the removal of potentially important parameters.
For fully connected and recurrent layers, instead of using the dense parameter $W$, the sparse product $W\circ M^p$ is used in the batched matrix multiplication, where $\circ$ represents the Hadamard product operator. As for convolutional layers, each convolution kernel is flattened to obtain $W$, following a process similar to that of fully connected layers.

Inspired by the dynamic sparse training \cite{liu2020dynamic}, we separate important and unimportant parameters by learning a threshold for each fully connected and convolutional neural network layer during training on one dataset. 
This threshold is a trainable parameter that is updated along with the backpropagation of the neural network to achieve a stepwise update. 
In order to make the binary step function $S(x)$ in threshold vector $t$ trainable via back-propagation, a derivative estimation is needed. A long-tailed higher-order estimator $H(x)$ proposed by \cite{xu2019accurate} is adopted for a balance of tight approximation and smooth back-propagation.
\begin{equation} \label{eq3}
\frac{d}{dx}S\left(x\right)\approx H\left(x\right)= \left\{
\begin{array}{ll}
      2-4|x| & -0.4, \leq x \leq 0.4 \\
      0.4, & 0.4<|x|\leq 1 \\
      0, & otherwise \\
\end{array} 
\right. 
\end{equation}
To get the pruning masks $M^p$ with high sparsity, higher pruning thresholds are needed. To achieve this, a sparse regularization term $L_s$ is added to the training loss that penalizes the low threshold value. For each trainable masked layer with threshold $t$, the corresponding regularization term is $R=\sum_{i=1}^{c_o}{exp(-t_i)}$. Thus, the sparse regularization term $L_s$ for a neural network with $C$ trainable masked layers is:
\begin{equation}\label{eq4}
L_s=\sum_{i=1}^{C}R_i
\end{equation}
$exp(-x)$ is used as the regularization function since it is asymptotical to zero as x increases. Consequently, it penalizes low thresholds without encouraging them to become extremely large.
Given the training dataset D, a sparse neural network can be trained directly with backpropagation algorithm by adding the sparse regularization term $L_s$ to the loss function as follows:
\begin{equation}\label{eq5}
W^\ast,t^\ast=argmin[L\left(D;W\right)+\alpha\ L_s]
\end{equation}
where $L\left(\cdot\right)$ is the loss function, e.g., cross-entropy loss for classification, and $\alpha$ is the scaling coefficient for the sparse regularization term, which can control the percentage of parameters remaining. It is calculated according to the total number of iterations of one dataset.
For a new dataset, the pruning threshold $t$ is re-initialized, and a new round of fine-grained data-driven pruning is restarted and be applied to neural network parameters not occupied by previous datasets.

\subsection{Task-Agnostic Parameter Isolation}

To address the issue of catastrophic forgetting with varying similarity datasets, we enhance the technique of parameter isolation. Traditional methods freeze learned parameters during both training and inference, preventing them from being updated and masking task-specific parameters. In contrast, our data-driven approach progressively learns from frozen parameters while utilizing all parameters during inference. This allows effective handling of heterogeneity in data similarity during continuous learning.
Parameter freezing in neural networks involves preventing specific parameters from being updated during training. In our approach, we introduce a binary freeze mask, denoted as $M$, of the same shape as the parameters. This mask has a value of 1 for parameters that are allowed to be updated and 0 for frozen parameters. We obtain the frozen parameters, $\theta_f$, by element-wise multiplying the original parameters by this mask:
$$
\theta_f = \theta \odot M
$$
During training, the gradients computed with respect to the loss function are applied only to the non-frozen parameters, updating them according to the optimization algorithm. The frozen parameters remain unchanged throughout the training process.
At the end of training on each dataset, we calculate a freeze mask $M^f$, which is the result of the union between the existing pruning mask and the freeze mask from the previous round. This mask is used to freeze the learned parameters during the next dataset training. 
The freeze mask $M^f$ is calculated as follows:
\begin{equation}\label{eq6}
{M^f}_{ij}=S\left(\left|{M^f}_{ij}+{M^p}_{ij}\right|\ \right),\ \ 1\le i\le c_o,1\le j\le c_i
\end{equation}
where $S$ is the sign function, $c_o$ is the number of output channels, $c_i$ is the number of input channels, and ${M^p}_{ij}$ is the pruning mask obtained after pruning. 

In order to ensure that the corresponding gradient of the parameters in the freeze mask ${M^f}_{ij}$ is set to $0$ when $W_{ij}$ needs to be frozen, we use the following equation:

\begin{equation}\label{eq7}
    W^*,t^*=argmin\left[L\left(D;W,t\right)+\alpha L_s\right]\circ (1-M^f)
\end{equation}
Here, $L(D;W,t)$ denotes the loss function on the current dataset, $L_s$ is the penalty for changes in learned parameters, $\alpha$ is the learning rate, $W^*$ and $t^*$ denotes the optimal value of the weight and threshold respectively, and $\circ$ denotes element-wise multiplication between the matrices. 
During inference, all the parameters, including the frozen ones, are used to make predictions, as the model has already learned useful representations from them. By applying parameter freezing, a neural network can retain knowledge from previous tasks while allowing for further learning without catastrophic forgetting.
For a new dataset, adaptive continual learning initiates a fresh iteration while preserving important frozen parameters. Pruning is only applied to other free neural network parameters. 

\subsection{Adaptive Continual Learning Training Flow}

\begin{algorithm}[t]
\caption{Training Flow of AdaptCL}
\begin{algorithmic}[1]
\STATE 	Require: weight of parameter $W$, threshold vector $t$ is initialized with zero tensor.
\FOR{dataset $d=\ 0,\ 1,\ 2,\ldots$ }
    \FOR{layer in model}
        \STATE 	Reset threshold $t\gets0$
    \ENDFOR
\FOR{epoch}
    \FOR{step}
        \STATE update pruning mask ${M^p}_{ij}=S\left(\left|W_{ij}\right|-t_i\ \right)$
        \STATE 	update pruned weight $W=W\circ M^p$
        \FOR{layer in model}
            \STATE 	update the loss $L\left(\cdot\right)=L\left(D;W\right)+\alpha L_s$
            \ENDFOR
        \IF{$d==0$}
            \STATE gradient decent $W^\ast,t^\ast=argminL\left(\cdot\right)$
        \ELSE
            \STATE 	gradient decent with frozen parameters $W^\ast,t^\ast=argminL\left(\cdot\right)\circ (1-M^f)$
            \ENDIF
        \ENDFOR
    \ENDFOR
    \STATE update freeze mask ${M^f}_{ij}=S\left(\left|{M^f}_{ij}+{M^p}_{ij}\right|\ \right)$
\ENDFOR
\end{algorithmic}
\label{algorithm1}
\end{algorithm}

Referring to the algorithm flow of our proposed method, depicted in Algorithm \ref{algorithm1}, at the start of each new round of dataset training, threshold parameters are initialized. During training, these threshold parameters are calculated and updated at each step of backpropagation, leading to the refinement of the pruning mask. 
$$
{M^p}_{ij}=S\left(\left|W_{ij}\right|-t_i\ \right)
$$ The refinement process, being fine-grained, data-driven, and step-wise, allows AdaptCL to adapt to variations in data complexity and dataset size.
Throughout the training process, AdaptCL freezes the gradient descent at each step based on the freeze mask $(1-M^f)$. This protects the current parameters from further modification, preserving the knowledge learned in previous training rounds and mitigating the impact of catastrophic forgetting caused by variations in data similarity.
At the end of each round of training, AdaptCL generates an updated freeze mask to protect the current set of parameters for future training. This allows AdaptCL to continue learning from new data while retaining the knowledge gained from previous training rounds.
Overall, the adaptive learning with fine-grained data-driven pruning approach, coupled with task-agnostic parameter isolation, enables AdaptCL to effectively adapt to variations in data complexity and dataset size while mitigating the impact of variation of data similarity during the training process.

\section{Experiments}

Our method is evaluated on a range of benchmark datasets with heterogeneous characteristics, encompassing various domains and tasks. 
To assess the performance of our method, we apply the method to the widely recognized ResNet-18, LeNet-5 and VGG-16 architectures. To establish a solid benchmark for comparison, we implement several other baseline algorithms in the domain incremental setting. These algorithms include SGD as the naive setting, as well as EWC, LwF, PRE-DFKD, PackNet*, and Separated Models for Learning (SML). Particularly, PackNet* represents an extension of PackNet specifically designed for our task-agnostic evaluation.

By conducting experiments on these benchmark datasets and comparing our method against these baseline algorithms, we aim to gain insights into the performance of our proposed approach and to assess its effectiveness in addressing the challenges of tackling data heterogeneity in continual learning.
In particular, we aim to answer the following research questions:
\begin{itemize}
\item \textbf{Q1}: How does AdaptCL compare to other baseline continual learning methods in terms of average accuracy and parameter efficiency?
\item \textbf{Q2}: What is the effectiveness of AdaptCL in managing heterogeneity in sequential datasets from different application domains, such as Food Quality and DomainNet?
\item \textbf{Q3}: What is the impact of AdaptCL's fine-grained data-driven pruning technique on adapting to differences in data complexity and dataset size?
\item \textbf{Q4}: How does AdaptCL's task-agnostic parameter isolation approach mitigate catastrophic forgetting in the presence of varying degrees of data similarity?

\end{itemize}

\subsection{Datasets}

We choose the following datasets to evaluate our method:

\subsubsection{\textbf{Large-Scale, Diverse Binary-Class Food Quality Dataset}}

\begin{figure}
  \begin{center}
\includegraphics[width=\linewidth]{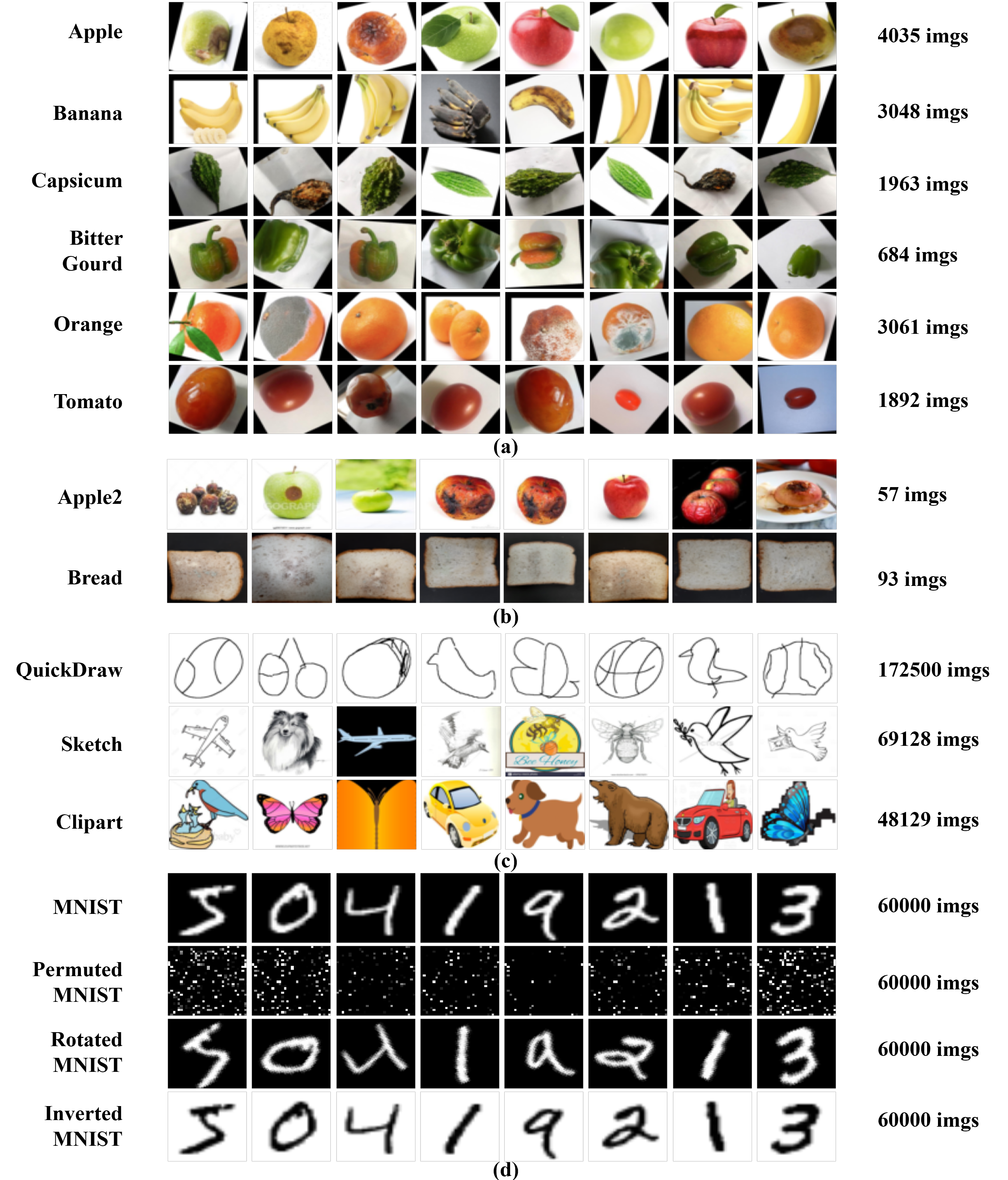}
  \\
  \caption{Examples of input images used in the experiments. (a) Large-Scale, Diverse Binary-Class Food Quality Dataset. (b) Few-Shot, Multi-Class Food Quality Dataset. (c) DomainNet comprises datasets with heterogeneous complexity and size.
  (d) MNIST Variants with heterogeneous similarity.}
 \label{figure3}
  \end{center}
\end{figure}

The dataset comprises a total of 14,683 images of six different types of fruits and vegetables, as shown in Figure \ref{figure3}(a), including apples, bananas, bitter gourds, capsicums, oranges, and tomatoes. Each image in the dataset is classified as either fresh or stale. The datasets vary in size, and the images are obtained from various sources such as online image repositories, self-captured images, or artificially-generated images through data augmentation techniques, resulting in different levels of complexity and similarity among the datasets. The datasets are designed to be heterogeneous and challenging to evaluate the robustness and generalization of machine learning models. 
All the images in the dataset have been preprocessed to ensure a uniform size and aspect ratio of $64\times64$ pixels. The total size of the dataset is approximately 2GB.

\subsubsection{\textbf{Few-Shot, Multi-Class Food Quality Dataset}}

The used as a real-life application case to verify our solution on a small dataset size, which poses a more challenging scenario compared to the previous binary classification dataset. This dataset comprises images of Apples and Bread, each associated with a freshness score label. The freshness scores range from 0 to 4, where 0 represents total corruption and 4 indicates total freshness. The Apple dataset consists of a total of 57 images, while the Bread dataset contains 93 images, as illustrated in Figure \ref{figure3}(b). This dataset aims to evaluate the model's performance in adapting to very few samples, and the ability to transfer knowledge to solve under-fitting.

\subsubsection{\textbf{DomainNet with heterogeneous complexity and size}}

The DomainNet\cite{peng2019moment} dataset consists of image data from six domains, each with a different amount of data, including real photos, painting, clipart, infograph, quickdraw, and sketch. There are 48K - 172K images (600K in total) categorized into 345 classes per domain.
DomainNet comprises datasets with heterogeneous complexity and size. For instance, the Quickdraw dataset holds 172,500 images but requires only 439MB of storage, while the Sketch dataset includes 69,128 images but occupies 2.5GB of storage.
The Sketch, Quickdraw, and Clipart domains are selected as datasets as shown in Figure \ref{figure3}(c) to evaluate models' performance on datasets with different complexity.

\subsubsection{\textbf{MNIST and variants with heterogeneous similarity}}

To provide additional validation for our model, we choose to utilize MNIST dataset with its variants, which include the MNIST, Permuted MNIST, Inverted MNIST, and Rotated MNIST datasets. These datasets are organized into sequences, reflecting a different level of similarity, and in both domain-incremental and class-incremental settings.

The datasets each consist of 70,000 images of handwritten digits from 0 to 9 of size 32 × 32. In each dataset, 60,000 images are used for training and 10,000 images for the test, as listed in Figure \ref{figure3}(d).

\begin{itemize}
\item \textbf{Permuted MNIST} is an MNIST variant that applies a fixed random permutation of the pixels of the MNIST digits. It also includes the same number of images of handwritten digits. Permuted MNIST bears no resemblance to MNIST at all.

 \item \textbf{Inverted MNIST} is another variant of the MNIST dataset inverting the color of MNIST images from black to white. The Inverted MNIST and MNIST are the exact opposite in the color of the input data and the same in the output.

 \item \textbf{Rotated MNIST} is also a variant of the MNIST dataset. It rotates MNIST data randomly by 0-45 degrees. There is some overlap of data between the MNIST and the Rotated MNIST, making the two datasets similar to each other.

\end{itemize}

\subsection{Networks Used}
To evaluate the applicability of our proposed technique on networks of different sizes and structures, we conducted our experiments using three popular network architectures: LeNet-5, ResNet-18, and VGG-16. 

LeNet-5 is a relatively simple architecture with 61,706 parameters and a compact size of 0.24 MB. It was primarily designed for digit recognition in checks and consists of 7 convolutional layers. However, due to its limited number of convolutional layers, LeNet-5 may face resource constraints when processing sequential datasets.

ResNet-18, on the other hand, is a more complex architecture with 11,172,810 parameters and a larger size of 42.62 MB. This network incorporates a greater number of convolutional layers, making it better equipped to handle complex image recognition tasks. The increased number of parameters allows for a larger network capacity, which is advantageous for continuous learning scenarios involving sequential datasets.

Lastly, we utilized the VGG-16 architecture, which is more parameter-rich, with 14,986,570 parameters and a size of 57.17 MB. This architecture offers a high degree of expressiveness due to its numerous convolutional and fully connected layers.

\subsection{Evaluation Metrics}
For a principled evaluation, we adopt the following evaluation metrics\cite{lopez2017gradient}:
\begin{itemize}
\item Average Accuracy:
$AAC=\frac{1}{T}\sum_{i=1}^{T}R_{T,i}$

\item Backward Transfer:
$BWT=\frac{1}{T-1}\sum_{i=1}^{T-1}\left(R_{T,i}-R_{i,i}\right)$

\item Forward Transfer:
$FWT=\frac{1}{T-1}\sum_{i=2}^{T-1}{R_{i-1,i}-\bar{b_i}}$
\end{itemize}

We consider access to a testing dataset for each of the D datasets. After the model finishes learning about the domain $t_i$, we evaluate its test performance on all $T$ datasets. By doing so, we construct the matrix $R\in R^{t\times t}$, where $R_{i,j}$ is the test classification accuracy of the model on the dataset $t_j$ after observing the last sample from dataset $t_i$. Letting $\bar{b}$ be the vector of test accuracy for each task at random initialization. 
For comparison, our primary criterion for evaluating performance is the average accuracy (AAC) metric, where higher values indicate better performance. Additionally, we consider the metrics of backward and forward transfer efficiency (BWT and FWT), with higher values being preferred. Furthermore, we calculate parameters (Params) to assess parameter efficiency. To gain a deeper understanding of model performance across datasets, we also compare test accuracy for each dataset.

\subsection{Baselines}
To validate the effectiveness of our method in continual learning with heterogeneous datasets, we compare our model with baseline algorithms.
We implement all of the following described baselines in our code base:

\begin{itemize}
    \item Separated model learning (SML): Separate models are trained for every task, achieving the highest possible accuracy by dedicating all the network resources to that single dataset. In this case, there is no knowledge transfer or catastrophic forgetting. It requires manual selection of the model during inference.

\item SGD\cite{bottou1991stochastic}: A naïve model trained with direct stochastic gradient descent.

\item EWC\cite{kirkpatrick2017overcoming}: A regularization technique in continual learning that uses diagonal elements of Fisher Information Matrix to constrain the weights of the neural network and avoid catastrophic forgetting.

\item LwF\cite{li2017learning}: A rehearsal-based method that uses knowledge distillation to preserve previously learned knowledge along with training on new tasks.

\item PRE-DFKD\cite{binici2022robust}: A recently proposed rehearsal strategy that rehearses the model using the data-free knowledge distillation through the distribution of the previously observed synthetic samples from a Variational Autoencoder.

\item PackNet\cite{mallya2018packnet}: A structure-based parameter isolation method that prunes a specific ratio of the network during training to sequentially "pack" multiple tasks into a single network. It requires knowing the number of datasets ahead to calculate the pruning ratio. Also, it needs to select masks to indicate network modules to perform during inference. We implement it in the task-agnostic setting referred to as PackNet* later in this paper.

\end{itemize}
\subsection{Implementation Details}

We use Pytorch and Torchvision libraries to implement neural networks. All of the training images are scaled and normalized before training as preprocessing. Identical processes are applied to the test images. The optimizer is stochastic gradient descent (SGD), with a 0.001 learning rate, 0.9 as the momentum value, and Nesterov Accelerated Gradient for regularization.
To guarantee completely reproducible results, we set seed value as 5 for the random function of Numpy, python Random, Pytorch, Pytorch Cuda, and set Pytorch backends Cudnn benchmark as False, with Deterministic as True, configuring PyTorch to avoid using nondeterministic algorithms for some operations, so that multiple calls to those operations, given the same inputs, will produce the same result.
Algorithm \ref{algorithm1} shows the learning procedure of AdaptCL.
We keep all the settings the same for our method and the baselines.

Considering the Fisher matrix of EWC, we use EWC $\lambda$ as $1$.
Regarding PackNet, we implement it in a domain-incremental setting, which we refer to as PackNet* in our paper. Instead of using a pre-trained model, we train it for the same number of epochs as other methods, selecting ten epochs of sparse training following pruning, as discussed in PackNet's paper. To ensure each dataset received equal attention, we prune the network to assign the same ratio of $1/T$ parameters per dataset, where $T$ is the number of datasets. 
For PRE-DFKD, we follow the default setting and use Kullback–Leibler Divergence (KLD) loss with a hyperparameter of $10^{-5}$.
Regarding LwF, we set the hyperparameters Alpha and Temperature to $1$ and $3$, respectively.
For the naïve settings with stochastic gradient descent (SGD), we simply fine-tune the network on each new dataset without making any network modifications.
For Separated model learning, we use one network for training on every single dataset and do not fine-tune it on other datasets.

To facilitate the reproducibility of our experiments, we have made the source \href{https://github.com/PepperJao/AdaptCL}{\textbf{code}} available.

\section{Results}

\subsection{Performance on Datasets with Varied Size (Q1,Q2,Q3)}

\subsubsection{Large-Scale, Diverse Binary-Class Food Quality Dataset}

\begin{table*}[htbp]
\caption{Performance evaluation of continual learning methods in terms of average accuracy (AAC), backward knowledge transfer (BWT), forward knowledge transfer (FWT), and number of used parameters on the Large-Scale, Diverse Binary-Class Food Quality Dataset.}
    \centering
    \begin{tabularx}{\linewidth}{c|XXXcXXXXXc}
        \toprule
      
&	\multirow{2}*{AAC↑} &	\multirow{2}*{BWT ↑} &	\multirow{2}*{FWT ↑} &	\multirow{2}*{Params ($\times10^7$) ↓} &\multicolumn{6}{c}{Test Accuracy↑}\\

\cmidrule{6-11}
 & & & &	&	Apple& Orange& Banana& Tomato &Gourd& Capsicum \\
\midrule
SML&	0.998&	-&	-&	6.704&	1.000&	0.995&	1.000&	0.995&	1.000&	1.000\\
\midrule
SGD&	0.650&	-0.419&	0.033&	1.117&	0.530&	0.425&	0.665&	0.425&	0.855&	1.000\\
LwF&	0.683&	-0.379&	0.072&	1.117&	0.640&	0.475&	0.550&	0.525&	0.905&	1.000\\
EWC&	0.727&	-0.326&	0.058&	1.117&	0.530&	0.610&	0.860&	0.485&	0.875&	1.000\\
PackNet*&	0.695&	-0.361&	0.049&	1.117&	0.56&	0.465&	0.735&	0.475&	0.935&	1.000\\
PRE-DFKD&	0.749&	-0.300&	0.085&	1.117&	0.570&	0.705&	0.775&	0.505&	0.940&	1.000\\
AdaptCL&	\textbf{0.782}&	\textbf{-0.252}&	0.041&	\textbf{1.091}&	\textbf{0.840}&	0.415&	\textbf{0.770}&	\textbf{0.705}&	\textbf{0.975}&	0.984\\
        \bottomrule
    \end{tabularx}%
  \label{table1}%
\end{table*}%

Our method achieves an average accuracy of 78.20\% on the six food datasets, surpassing baseline methods by 4.32\%. It outperforms other approaches in terms of final accuracy and has the lowest parameter count ($1.091\times 10^7$) while effectively overcoming catastrophic forgetting. Despite having fewer parameters, our model successfully fits up to six datasets, with some minor accuracy gaps compared to SML due to data complexity and the need for increased capacity. To unlock its full potential, we recommend scaling up the model for improved accuracy in continual learning.

\begin{figure}
  \begin{center}
 \includegraphics[width=\linewidth]{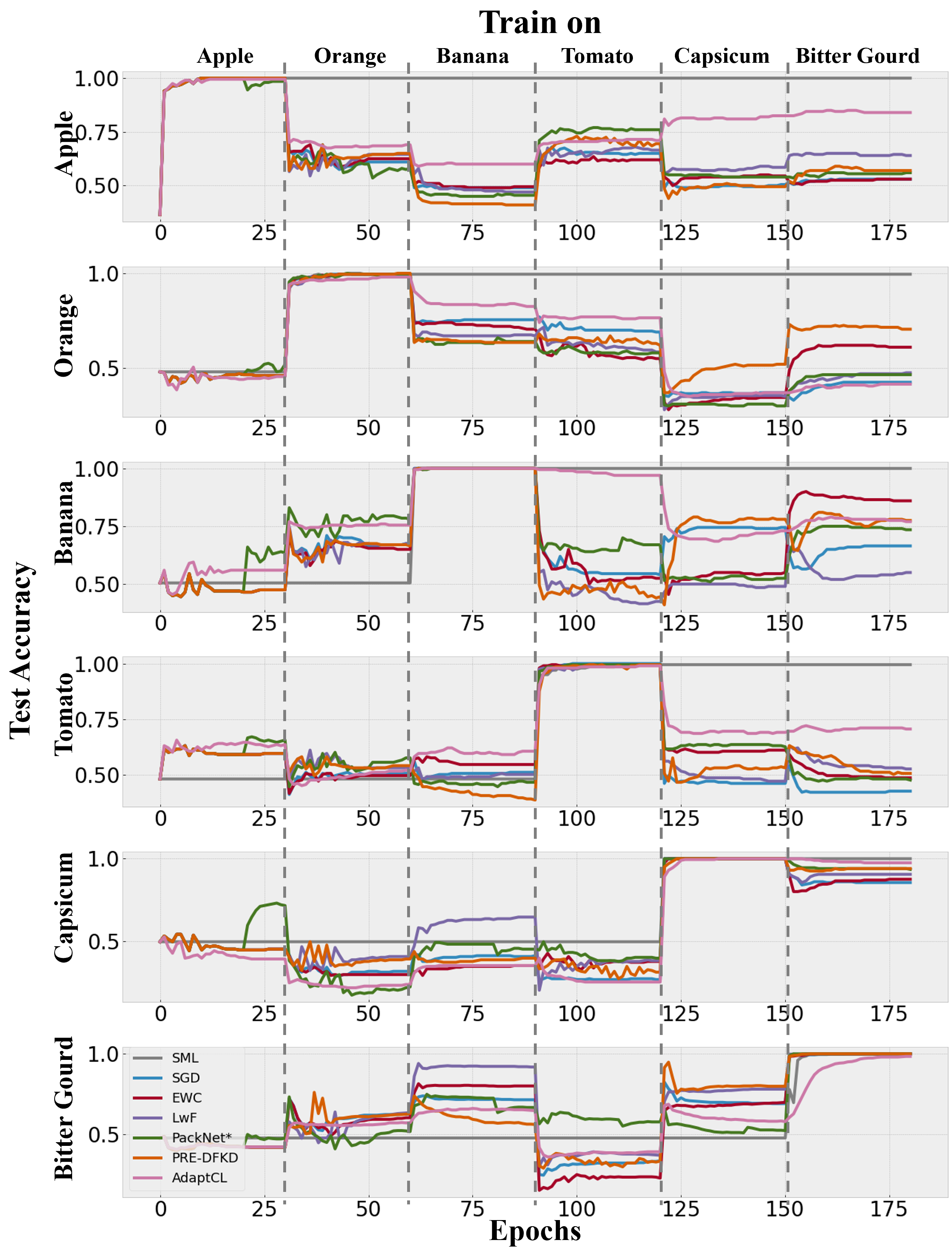}
  \\
  \caption{Test accuracy comparison of continual learning methods on the Large-Scale, Diverse Binary-Class Food Quality Dataset. Our proposed method, AdaptCL, achieves higher average accuracy while consistently preventing catastrophic forgetting in real-world applications with heterogeneous data, outperforming other methods. (Best viewed in color)}
 \label{figure4}
  \end{center}
\end{figure}

Analyzing Figure \ref{figure4}, we observe the varying impact of learning across datasets due to their heterogeneity. In most cases, our model maintains the highest accuracy on the learned datasets. Comparing it to PackNet*, which also uses pruning methods, we notice a notable accuracy increase on unlearned datasets during pruned epochs, indicating the efficacy of pruning for enhancing generalization in continual learning with relevant datasets.
Our model struggles with the Orange dataset following training on Capsicum and Bitter Gourd datasets due to conflicting features, mainly caused by the low data resolution of $64 \times 64$ pixels, which led the model to primarily rely on color and shape to differentiate images. This issue, observed in all baseline models, can be resolved by increasing data resolution.

\subsubsection{Few-Shot, Multi-Class Food Quality Dataset}

We test our method on this Dataset to evaluate its ability to generalize on small datasets. Like the human brain, which excels at few-shot learning and generalizing from limited examples, our AdaptCL method, inspired by the neural reuse principle, improves efficiency and performance on small sample datasets.
As shown in Table \ref{table2}, when evaluating the Few-Shot, Multi-Class Food Quality Dataset, AdaptCL achieves 99.50\% accuracy using 10\% fewer parameters than baseline methods, even producing a rare positive backward knowledge transfer of \textbf{1.08\%}, meaning the positive consequence of inductive knowledge transfer is more significant than catastrophic forgetting. Since the first dataset is small, the model is not fully trained, and easy to overfit; the new dataset can make the network more robust to have higher accuracy during inference on the test dataset.
Our model outperforms the baselines' AAC by 11.20\% and is superior to using a separated model for learning (SML) on the Few-Shot, Multi-Class Food Quality Dataset sequence with only 45\% of SML's parameters. These results demonstrate the potential advantages of our model when encountering a continuous stream of smaller datasets.

\begin{table*}[htbp]
\caption{Comparison of average accuracy (AAC), backward knowledge transfer (BWT), forward knowledge transfer (FWT), and the number of used parameters of various continual learning methods on the Few-Shot, Multi-Class Food Quality Dataset with different training orders.}
    \centering
    \begin{tabularx}{\linewidth}{c|cccccc|cccccc}
        \toprule
        &\multicolumn{6}{c|}{\textbf{Bread → Apple2}}&	\multicolumn{6}{c}{\textbf{Apple2 → Bread}}\\
        \midrule
    &	\multirow{2}*{AAC↑}&	\multirow{2}*{BWT↑}&	\multirow{2}*{FWT↑}&	\multirow{2}*{Params($\times10^7$)↓}&	\multicolumn{2}{c|}{Test Accuracy↑} &	\multirow{2}*{AAC↑}&	\multirow{2}*{BWT↑}&	\multirow{2}*{FWT↑}&	\multirow{2}*{Params($\times10^7$)↓}&	\multicolumn{2}{c}{Test Accuracy↑} \\
      \cmidrule{6-7}  \cmidrule{12-13} &&&&&Bread&Apple&&&&&Apple&Bread

    \\
    \midrule
SML&	0.989&	 - &	 - &	2.235&	0.978&	1.000&	0.989&	-&	-&	2.235&	1.000&	0.978\\
\midrule
SGD&	0.774&	-0.430&	0.368&	1.117&	0.548&	1.000&	0.849&	-0.281&	0.452&	1.117&	0.719&	0.978\\
LwF&	0.782&	-0.398&	0.368&	1.117&	0.581&	0.982&	0.849&	-0.281&	0.452&	1.117&	0.719&	0.978\\
EWC&	0.763&	-0.452&	0.368&	1.117&	0.527&	1.000&	0.831&	-0.316&	0.452&	1.117&	0.684&	0.978\\
PackNet*&	0.894&	-0.172&	0.281&	1.117&	0.806&	0.982&	0.893&	-0.193&	0.398&	1.117&	0.807&	0.978\\
PRE-DFKD&	0.859&	-0.086&	0.404&	1.117&	0.892&	0.825&	0.867&	-0.158&	0.409&	1.117&	0.842&	0.892\\
AdaptCL&	\textbf{0.995}&	\textbf{0.011}&	0.316&	\textbf{1.014}&	\textbf{0.989}&	\textbf{1.000}&\textbf{0.980}&	\textbf{-0.018}&	0.441&	\textbf{1.005}&	\textbf{0.982}&	\textbf{0.978}\\

        \bottomrule
    \end{tabularx}%
    \label{table2}%
\end{table*}%

\textbf{Effect of Training Orders}
To investigate the impact of training orders, we conduct experiments using different dataset sequences. Our method, AdaptCL, consistently achieves the best results in both forward and reverse training orders, as shown in Table \ref{table2}. Unlike baseline methods, the average accuracy (AAC) of AdaptCL remains unaffected by the training sequence, while the final accuracy of methods like SGD, LwF, and EWC is heavily influenced by the order of training.
This can be attributed to AdaptCL's fine-grained pruning and task-agnostic parameter isolation, which minimize catastrophic forgetting and promote model generalization, enabling adaptation to new datasets regardless of their presentation order. These findings demonstrate the significant impact of training order on the performance of traditional methods, likely due to the tendency to overfit early datasets during training.
The robustness of AdaptCL to training order positions it as a preferred method for domains requiring frequent learning and adaptation to new datasets, as it effectively avoids the limitations associated with traditional methods.

\subsection{Performance on Datasets with Varied Complexity (Q1,Q2,Q3)}

\begin{table*}[htbp]
\caption{Results of different continual learning methods on the DomainNet dataset, including their AAC, BWT, FWT, and the number of used parameters. Our proposed method, AdaptCL, demonstrates the best AAC and BWT, indicating its ability to handle datasets with heterogeneous dataset size and complexity.}
    \centering
    \begin{tabularx}{6in}{c|XXXcXXX}
        \toprule
        
&	\multirow{2}*{AAC ↑}&	\multirow{2}*{BWT ↑}&	\multirow{2}*{FWT ↑}&	\multirow{2}*{Params($\times10^7$) ↓}&\multicolumn{3}{c}{Test Accuracy↑}	\\
\cmidrule{6-8}
&	&	&	&	&Quickdraw&Sketch	&Clipart\\
\midrule 
SML&	0.624&	-&	-&	3.404&	0.774&	0.573&	0.524\\
\midrule
SGD&	0.449&	-0.220&	0.209&	1.135&	0.394&	0.414&	0.539\\
LwF&	0.448&	-0.237&	0.103&	1.135&	0.381&	0.411&	0.552\\
EWC&	0.457&	-0.203&	0.103&	1.135&	0.421&	0.414&	0.536\\
PackNet*&	0.484&	-0.180&	0.098&	1.135&	0.483&	0.448&	0.521\\
PRE-DFKD&	0.461&	-0.206&	0.109&	1.135&	0.464&	0.416&	0.504\\
AdaptCL&	\textbf{0.531}&	\textbf{-0.131}&	0.106&	\textbf{1.051}&	\textbf{0.570}&	\textbf{0.510}&	0.512\\

        \bottomrule
    \end{tabularx}%
  \label{table3}%
\end{table*}%

\begin{figure}
  \begin{center}

{\includegraphics[width=\linewidth]{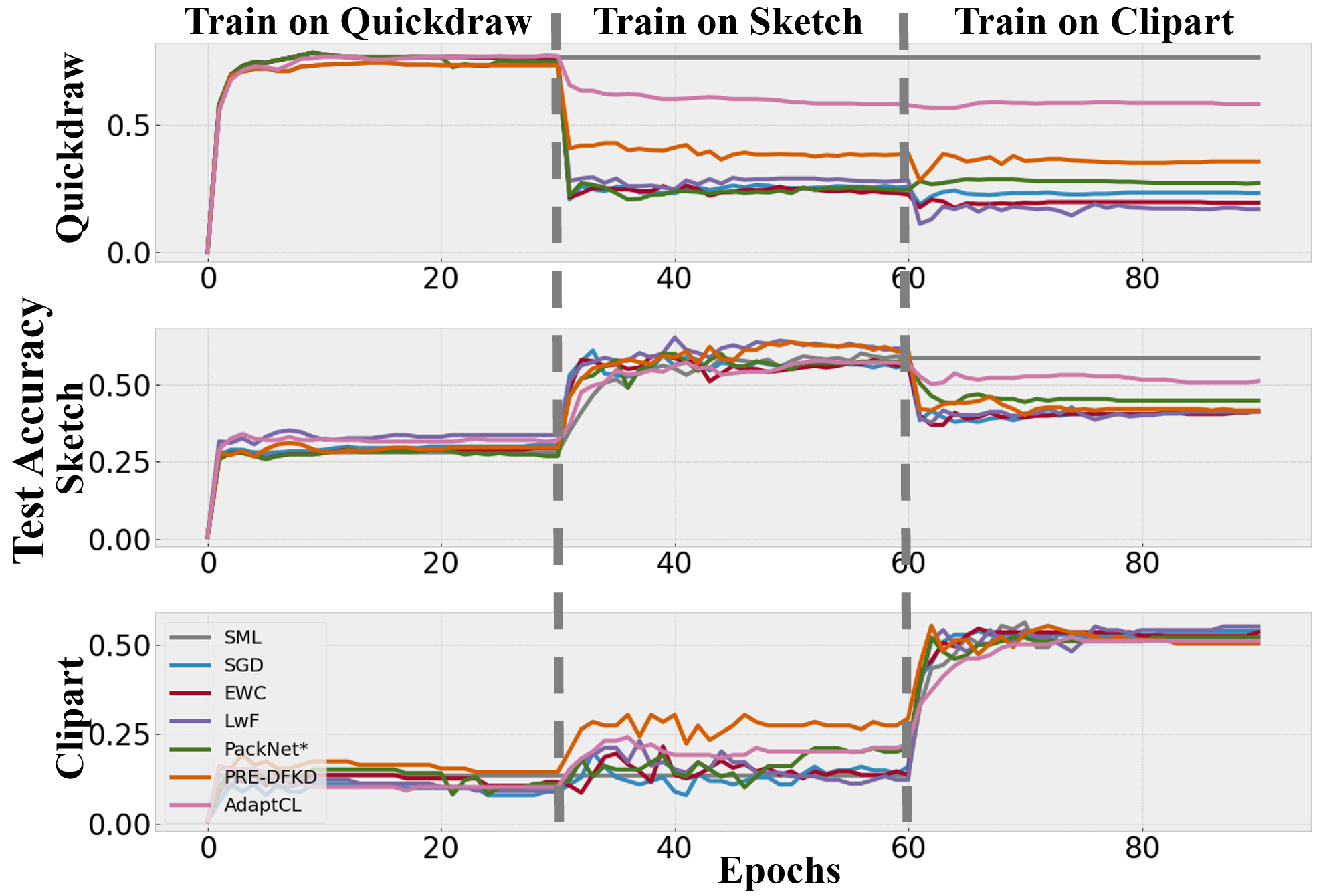}}
 \caption{Results of continual learning methods on the DomainNet that comprises datasets with heterogeneous complexity and size. AdaptCL achieves the best average accuracy and is the most robust to datasets with varied complexity and size. (Best viewed in color)}
  \label{figure5}
  \end{center}
\end{figure}

We evaluate the performance of AdaptCL on the DomainNet sequence, which is a heterogeneous classification dataset made up of images from different domains, each of varying size and complexity \cite{peng2019moment}. 
On the DomainNet dataset, rehearsal-based methods like LwF and PRE-DFKD perform poorly, especially LwF, even lower than SGD without any Continual Learning method assistance. This is likely due to the large and disparate sizes of the three subsets in DomainNet, making it challenging to adjust simple knowledge distillation methods based on dataset size.
Additionally, comparing SML with other methods on the last subset, Clipart, we observe that learning models on sequential datasets can facilitate faster learning and forward knowledge transfer, resulting in higher test accuracy compared to separated model learning (SML). The AdaptCL doesn't achieve higher accuracy than SML on this subset due to pruning, which makes the model more parameter efficient, but simultaneously slows down the learning of new data because of insufficient model capacity. This issue can be solved by network expansion.
As shown in Table \ref{table4}, AdaptCL outperforms the baselines, improving network performance by 18.24\% in average accuracy and 44.79\% in backward transfer compared to SGD, while beating the baselines CL methods by 9.70\% in AAC and 30.69\% in BWT, by using only 92.65\% of their parameters. Despite the impressive results, gaps in accuracy persisted compared to using separate models for learning, primarily due to the complex nature of the DomainNet data that demand increased model capacity to handle more complex information with significant distribution shifts within the dataset.
From Figure \ref{figure5}, we can see that even with significant differences in data, Clipart, Sketch, and Quickdraw can rely on forward knowledge transfer to achieve faster learning. In the context of continual learning, old datasets can improve the accuracy of new datasets, making CL methods more accurate than using separate models for learning (SML) to learn new data. Among the methods evaluated, rehearsal is the most effective in promoting faster learning, while AdaptCL excelled in accuracy retention.

\subsection{Performance on Datasets with Varied Similarity (Q1,Q4)}

\subsubsection{Dissimilar MNIST Variants}

\begin{table*}[htbp]
\caption{Comparison of average accuracy (AAC), backward knowledge transfer (BWT), forward knowledge transfer (FWT), and the number of used parameters of different continual learning methods on the dissimilar MNIST Variants dataset.}
    \centering
    \begin{tabularx}{6in}{c|XXXcccc}
        \toprule
        
&	\multirow{2}*{AAC↑}&
\multirow{2}*{BWT↑}&	\multirow{2}*{FWT↑}&	\multirow{2}*{Params ($\times10^7$)↓}&	\multicolumn{3}{c}{Test Accuracy↑}\\
\cmidrule{6-8}
&	&	 &	 & &	MNIST&	Permuted MNIST&	Inverted MNIST\\
\midrule
SML&	0.989&	-&	-&	3.352&	0.993&	0.980&	0.993\\
\midrule
SGD&	0.755&	-0.351&	0.006&	1.117&	0.483&	0.787&	0.994\\
LwF&	0.643&	-0.521&	-0.002&	1.117&	0.331&	0.604&	0.994\\
EWC&	0.753&	-0.354&	0.009&	1.117&	0.363&	0.776&	0.993\\
PackNet*&	0.841&	-0.222&	0.009&	1.117&	0.591&	0.939&	0.993\\
PRE-DFKD&	0.784&	-0.274&	0.011&	1.117&	0.723&	0.686&	0.943\\
AdaptCL&	\textbf{0.967}&	\textbf{-0.019}&	\textbf{0.023}&\textbf{1.046}&	\textbf{0.980}&	\textbf{0.936}&	\textbf{0.986}\\
        \bottomrule
    \end{tabularx}%
  \label{table4}%
\end{table*}%

We screen two sets of MNIST Variants to compose similar and dissimilar sequences, with the dissimilar sequence as MNIST, Permuted MNIST, and Inverted MNIST.
Regarding the dissimilar MNIST Variant sequence, AdaptCL significantly improves the network's average accuracy (AAC) by 28.14\% and alleviates forgetting (BWT) by 94.50\% (Table \ref{table3}). It outperforms baselines by 15.03\% in AAC and 91.30\% in BWT on this sequence. Compared to separated model learning (SML) where separate models are trained for each task, AdaptCL achieves comparable AAC while utilizing only 31.20\% of SML's parameters. Our method’s ability to minimize forgetting while learning dissimilar datasets, its parameter efficiency, and generalization contribute to its effectiveness.

\begin{figure}
  \begin{center} {\includegraphics[width=\linewidth]{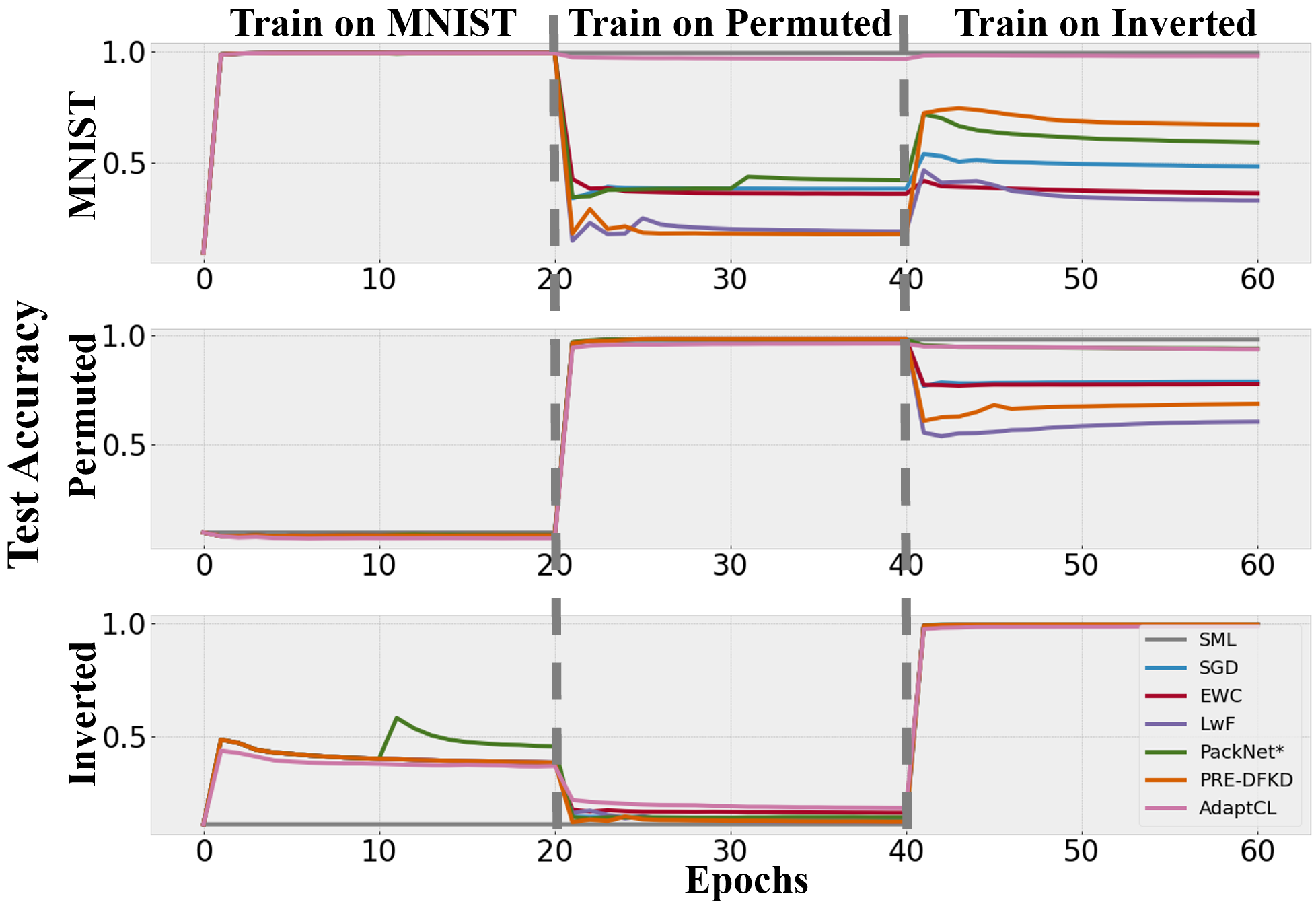}}
 
  \caption{Test accuracy comparison of continual learning methods on three dissimilar MNIST Variant datasets. Compared with using separated models for learning (SML), AdaptCL achieved comparable AAC results while using only 31.20\% of SML's parameters. AdaptCL's ability to achieve minimal forgetting while learning dissimilar datasets, coupled with its parameter efficiency, establishes the effectiveness of our approach. (Best viewed in color)}
  \label{figure6}
  \end{center}
\end{figure}

Rehearsal and regularization-based methods like EWC, LwF, and PRE-DFKD perform poorly on the Dissimilar MNIST Variants dataset due to the vast data amount and dissimilarity between datasets. Parameter isolation-based methods like PackNet* and our method, AdaptCL, demonstrate significant advantages on this dataset. Training with plain SGD leads to catastrophic forgetting and a performance decline of at least 50\% (Figure \ref{figure6}). EWC slows down the performance decline initially, but it deteriorates over time. While PackNet* shows some improvement through pruning during learning on Dataset A, it is not as effective as AdaptCL in inhibiting catastrophic forgetting. AdaptCL, with a fixed neural network, adapts to new datasets while maintaining high performance on previous datasets without significant forgetting.

\subsubsection{More similar MNIST Variants}

\begin{table*}[htbp]
\caption{Comparison of average accuracy (AAC), backward knowledge transfer (BWT), forward knowledge transfer (FWT), and the number of used parameters of different continual learning methods on more similar MNIST Variant datasets. }
    \centering
    \begin{tabularx}{6in}{c|XXXcccc}
        \toprule
&	\multirow{2}*{AAC↑}&	\multirow{2}*{BWT↑}&	\multirow{2}*{FWT↑}&	\multirow{2}*{Params ($\times10^7$)↓}&	\multicolumn{3}{c}{Test Accuracy↑}\\
\cmidrule{6-8}
&	 & & & &	MNIST&Permuted MNIST&	Rotated MNIST\\
\midrule
SML&	0.989&	-&	-&	3.352&	0.993&	0.980
&0.992\\
\midrule

SGD&	0.884&	-0.156&	0.077&	1.117&	0.994&	0.666&	0.993\\
LwF&	0.729&	-0.391&	0.023&	1.117&	0.993&	0.204&	0.991\\
EWC&	0.889&	-0.149&	0.088&	1.117&	0.994&	0.681&	0.992\\
PackNet*&	0.938&	-0.076&	0.101&	1.117&	0.994&	0.830&	0.991\\
PRE-DFKD&	0.822&	-0.240&	0.179&	1.117&	0.991&	0.488&	0.988\\
AdaptCL&	\textbf{0.958}&	\textbf{-0.032}&	\textbf{0.339}&	\textbf{1.044}&	0.990&	\textbf{0.900}&	0.985\\

        \bottomrule
    \end{tabularx}%
    \label{table5}%
\end{table*}%

\begin{table*}
\caption{Comparison of average accuracy (AAC), backward knowledge transfer (BWT), forward knowledge transfer (FWT), and the number of used parameters of different continual learning methods in the class-incremental setting.
}
    \centering
    \begin{tabularx}{6in}{c|XXXcccc}
        \toprule
        
&	\multirow{2}*{AAC↑}&
\multirow{2}*{BWT↑}&	\multirow{2}*{FWT↑}&	\multirow{2}*{Params ($\times10^7$)↓}&	\multicolumn{3}{c}{Test Accuracy↑}\\
\cmidrule{6-8}
&	&	 &	 & & Class 0,1,2,3&	Class 2,3,4,5,6&	Class 0,4,7,8,9\\
\midrule
SML&	0.996&	-&	-&	3.352&	0.998&	0.995&	0.994\\
\midrule
SGD&	0.475&	-0.781&	0.208&	1.117&	0.231&	0.199&	0.995\\
LwF&	0.476&	-0.781&	0.208&	1.117&	0.234&	0.199&	0.995\\
EWC&	0.477&	-0.780&	0.208&	1.117&	0.208&	0.208&	0.996\\
PackNet*&	0.475&	-0.781&	0.207&	1.117&	0.231&	0.200&	0.995\\
PRE-DFKD&	0.476&	-0.780&	0.198&	1.117&	0.234&	0.199&	0.996\\
AdaptCL&	\textbf{0.652}&	\textbf{-0.508}&	0.205&	\textbf{1.089}&	\textbf{0.429}&	\textbf{0.543}&	0.983\\
        \bottomrule
    \end{tabularx}%
  \label{table8}%
\end{table*}%

\begin{figure}
  \begin{center} {\includegraphics[width=\linewidth]{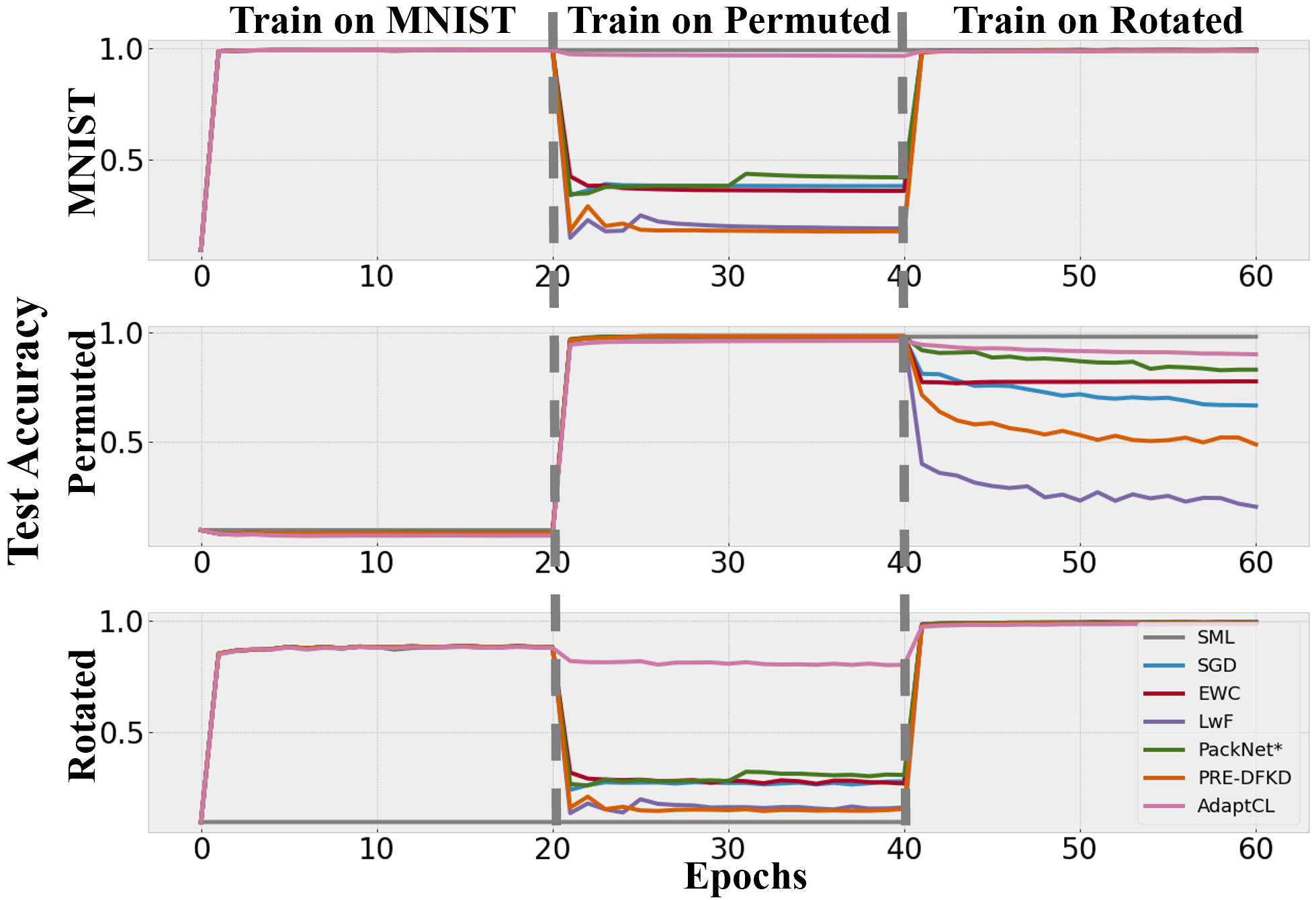}}
 
  \caption{Visualization of each method's test accuracy training on more similar MNIST Variant datasets. Still, AdaptCL retains the best performance compared with other CL methods. Compared with using separated models for learning (SML), AdaptCL can increase the accuracy on similar unseen datasets via forward knowledge transfer. (Best viewed in color)}
  \label{figure7}
  \end{center}
\end{figure}

In a similar MNIST Variant sequence consisting of MNIST, Permuted MNIST, and Rotated MNIST, our method maintains the highest accuracy, outperforming baselines by 21.80\% while using fewer parameters (Table \ref{table5}). AdaptCL achieves significant forward knowledge transfer on similar datasets with minimal catastrophic forgetting. Although it may not achieve the best accuracy on particularly similar datasets like MNIST and Rotated MNIST, our model consistently performs well on datasets with heterogeneous similarity, avoiding overfitting to similar datasets.
As dataset similarity increases, EWC suppresses catastrophic forgetting and achieves higher average accuracy compared to SGD, which differs from results on dissimilar datasets. LwF and PRE-DFKD struggle to balance dataset importance, leading to significantly lower accuracy on Permuted MNIST, indicating their unsuitability for large datasets with varying similarities. Our method demonstrates robust performance in such scenarios.

In Figure \ref{figure7}, all continual learning methods' inference accuracies on Rotated MNIST initially rise due to its similarity to MNIST. AdaptCL minimizes catastrophic forgetting and maintains high accuracy and generalization. Other CL methods, compared to separated model learning (SML), enhance model generalization, but still experience catastrophic forgetting while learning on Permuted MNIST. Notably, rehearsal-based methods like LwF and PRE-DFKD exacerbate catastrophic forgetting on Permuted MNIST when learning Rotated MNIST due to their similarity to the initial MNIST dataset.

\subsection{Performance in Class-Incremental Setting (Q1,Q3,Q4)}

The methods are evaluated in challenging class-incremental learning settings, with heterogeneous classes as shown in Table \ref{table8}. These datasets included class overlap and variations in the number of classes, introducing differences in data similarity and dataset size.
This setting poses a significant challenge for continual learning, as similar inputs leading to distinct classes that are assigned to different output layers, often causing rapid forgetting of previously learned classes within a single epoch.

Due to high intensity of catastrophic forgetting and large dataset sizes, techniques like rehearsal and regularization methods struggle to maintain accuracy effectively, as demonstrated in Table \ref{table8}. Their performance is only marginally superior to SGD.
Notably, when employing equal-proportional pruning, PackNet* exhibits poorer results in class-incremental settings in comparison to domain-incremental ones. This discrepancy is attributed to the significant impact of the pruning ratio on accuracy retention in class-incremental settings. In this context, catastrophic forgetting is prevalent across all methods and notably higher than in the domain-incremental setting, even with a reduced number of training epochs.
AdaptCL, on the other hand, stands out by significantly preserving accuracy, achieving an impressive 37.27\% higher average accuracy and reducing forgetting by 34.96\% when compared to SGD. This strong performance underscores AdaptCL's potential in mitigating catastrophic forgetting in task-agnostic settings.

\begin{table*}[htbp]
\caption{Results of different continual learning methods applied on LeNet-5 and VGG-16, including their AAC, BWT, FWT, and the number of used parameters on datasets (D1) MNIST (D2) Permuted MNIST, and (D3) Inverted MNIST.}
    \centering
    \begin{tabularx}{\linewidth}{c|ccccccc|ccccccc}
        \toprule
        &\multicolumn{7}{c|}{\textbf{LeNet-5}}&	\multicolumn{7}{c}{\textbf{VGG-16}}\\
        \midrule
    &	\multirow{2}*{AAC↑}&	\multirow{2}*{BWT↑}&	\multirow{2}*{FWT↑}&	Params&	\multicolumn{3}{c|}{Test Accuracy↑} &	\multirow{2}*{AAC↑}&	\multirow{2}*{BWT↑}&	\multirow{2}*{FWT↑}&	Params&	\multicolumn{3}{c}{Test Accuracy↑} \\
      \cmidrule{6-8}  \cmidrule{13-15} &&&&($\times10^4$)↓&D1&D2&D3&&&&($\times10^7$)↓&D1&D2&D3

    \\
    \midrule
SML&	0.968&-	&- &	18.51&	0.948&	0.978&	0.977&	0.990&	-&	-&	4.496&	0.995&	0.980&	0.995\\
\midrule
SGD&	0.479&	-0.740&	0.049&	6.171&	0.233&	0.220&	0.983&0.730&	-0.390&	-0.004&	1.499&	0.441&	0.753&	0.994\\
LwF&	0.542&	-0.646&	0.045&	6.171&	0.243&	0.398&	0.984&	0.763&	-0.342&	-0.342&	1.499&	0.532&	0.762&	0.995\\
EWC&	0.098&	-0.839&	0.003&	6.171&	0.098&	0.098&	0.098&0.747&	-0.363&	0.021&	1.499&	0.455&	0.792&	0.994\\
PackNet*&	0.533&	-0.644&	-0.018&	6.171&	0.141&	0.482&	0.975&	0.802&	-0.281&	-0.015&	1.499&	0.527&	0.885&	0.885\\
PRE-DFKD&	0.579&	-0.479&	0.027&	6.171&	0.421&	0.332&	0.985&	0.865&	-0.182&	0.010&	1.499&	0.841&	0.767&	0.989\\
AdaptCL&	0.567&	\textbf{-0.457}&	0.041&	\textbf{6.162}&	0.220&	\textbf{0.740}&	0.742&	\textbf{0.963}&	\textbf{-0.027}&	-0.003&	\textbf{1.396}&	\textbf{0.979}&	\textbf{0.922}&	0.987\\

        \bottomrule
    \end{tabularx}%
    \label{table7}%
\end{table*}%

\subsection{Performance on Different Networks (Q1)}

AdaptCL can be applied to neural networks with fully connected, recurrent, or convolutional layers. We also apply our method to the shallow Lenet-5 network and more complex VGG-16 to test its performance over different networks. 
For Lenet-5 (Table \ref{table7}), we observed a significant reduction in catastrophic forgetting compared to ResNet-18 when using AdaptCL. It achieved the second-highest AAC and the best BWT, trailing only PRE-DFKD, likely due to its lower parameter usage. Even with limited network capacity, AdaptCL outperformed other methods in mitigating catastrophic forgetting. Rehearsal-based approaches also showed promise in cases of insufficient model parameters and capacity.

On VGG-16 (Table \ref{table7}), AdaptCL exhibited its effectiveness on MNIST Variants datasets, surpassing baseline methods with the best average accuracy for AAC and BWT. Notably, most continual learning methods yielded negative FWT when applied to VGG-16. In the case of Permuted MNIST, VGG-16's performance did not improve after learning the MNIST dataset, possibly due to dissimilar image structures and VGG-16's inability to recognize permutations, whereas ResNet-18 performed slightly better in this scenario.

\subsection{Ablation Study (Q3,Q4)}

\subsubsection{Parameter Isolation Ratio}
\begin{figure}
  \begin{center}
  \subfigure[]{\includegraphics[width=1.7in]{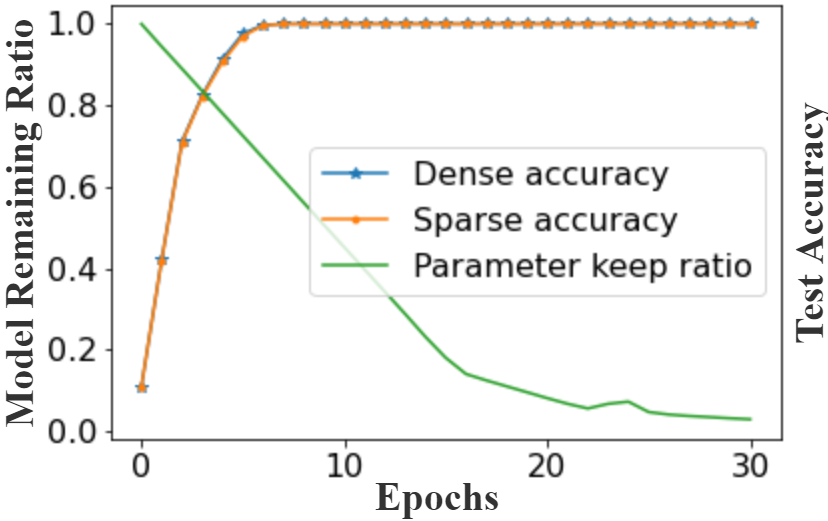}}
  \subfigure[]{\includegraphics[width=1.7in]{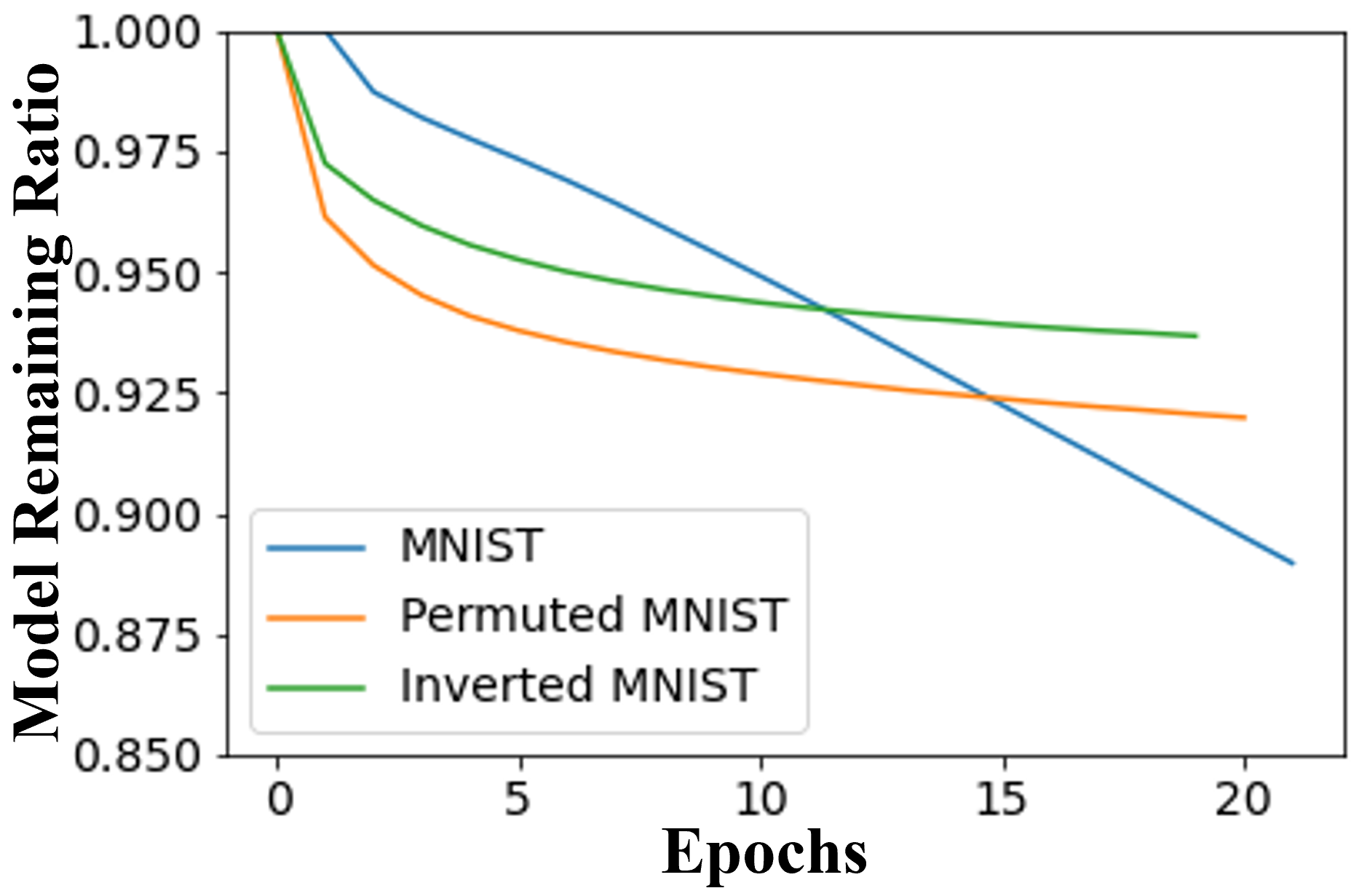}}
  \\
  \caption{(a) Model on ResNet-18 remaining ratio and sparse accuracy compared with dense accuracy, using $\alpha=10^{-4}$. (b) Change of model remaining ratio during training on MNIST Variants. (Best viewed in color)} \label{figure8}
  \end{center}
  
\end{figure}

To validate our intuition, we analyze and visualize the pruning ratio of the model in different datasets. Figure \ref{figure8}(a) displays the epoch-wise accuracy changes of the sparse network compared to fine-grained data-driven pruning and the dense network without pruning, along with the corresponding model remaining ratio during training.
Our method dynamically and adaptively learns the model remaining ratios during training on each dataset, rather than manually setting fixed ratios as in other pruning methods. Additionally, fine-grained data-driven pruning enables a highly sparse pruned network to achieve the same accuracy as a dense network.
Further, Figure \ref{figure8}(b) illustrates the change in the remaining ratios of the ResNet-18 model for each dataset of MNIST Variants at each epoch.
Figure \ref{figure8}(b) demonstrates that it is possible to fit the new dataset without sacrificing the accuracy of the old dataset by adding only a few parameters, even when there are significant differences in data distribution between the old and new datasets. Consequently, manually assigning the same parameter ratio to all datasets is not reasonable.


\subsubsection{Parameter Execution Pattern}
\begin{figure}
  \begin{center}
  \subfigure[]{\includegraphics[width=\linewidth]{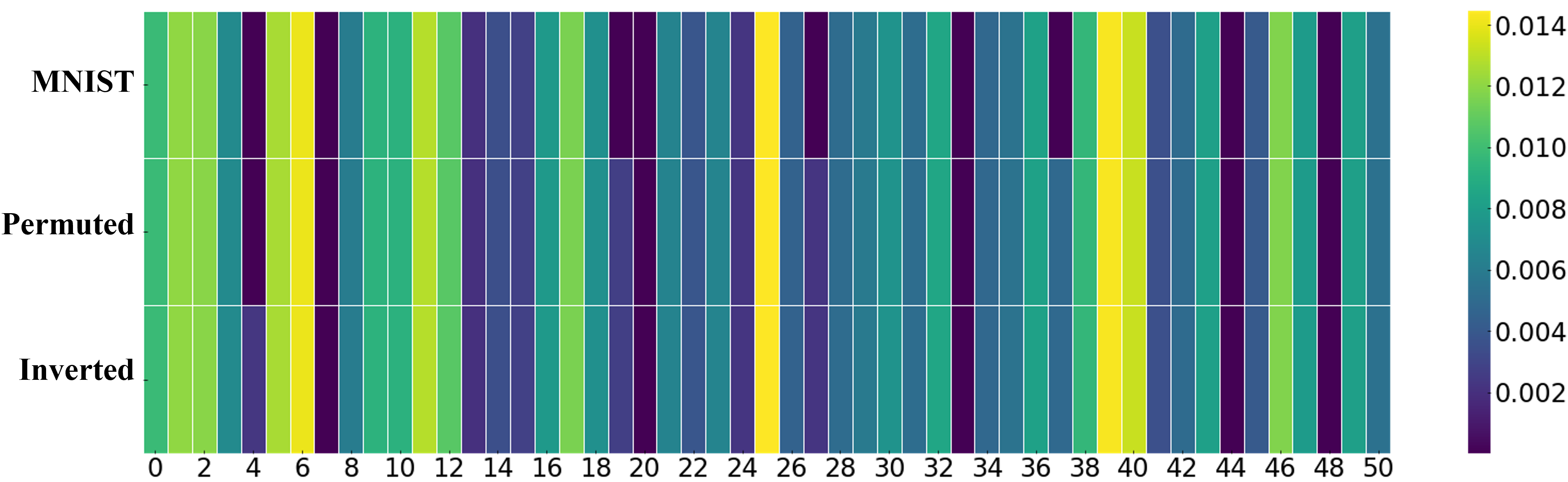}}
  \subfigure[]{\includegraphics[width=\linewidth]{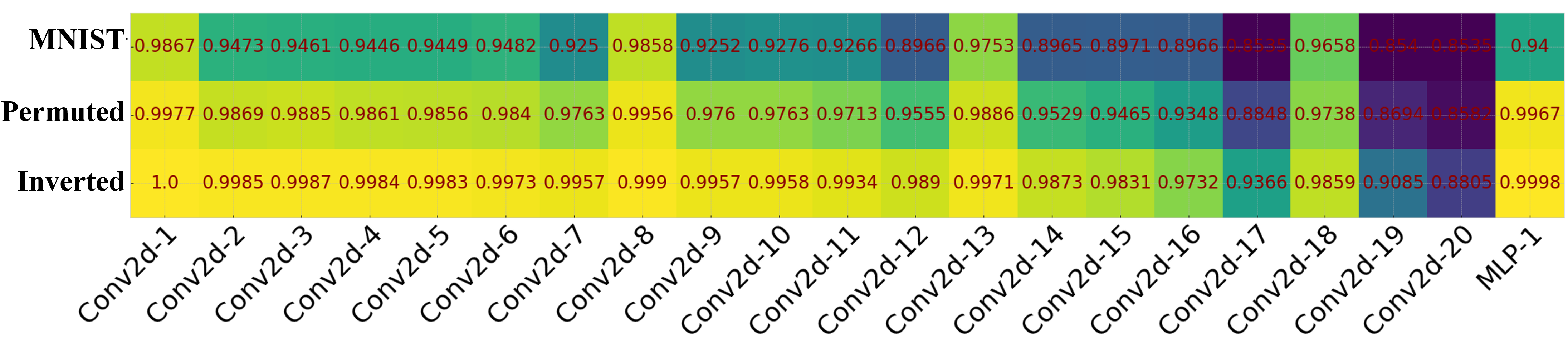}}
  \\
  \caption{Illustration of the parameter execution patterns for continually trained models on MNIST Variants. Heatmap (a) showcases the firing probability of the $x$th parameters within the 1st Conv2d layer demonstrated in the x-axis; it illustrates that subsequent datasets maintain and reuse the previous parameters and generalize by adding new connections to them. The heatmap (b) shows the utilization ratio of different layers.} \label{figure9}
  \end{center}
\end{figure}

To analyze the parameter reuse in AdaptCL, we visualize the pattern of parameter execution during training and the proportion of each layer in the neural network occupied progressively by different datasets. 
Figure \ref{figure9}(a) displays the pattern of parametric ignition of the first Conv2d layer (flattened) on the MNIST Variants. Figure \ref{figure9}(b) presents the proportion of parameters occupied during adaptive learning in each convolution and fully connected layer of ResNet-18.
From these figures, we observe that the parameters activated during previous task training remain unchanged when learning a new dataset. Additionally, we notice that some previously unused parameters (black) become activated during the new dataset training, contributing to the overall generalization of the network.

\subsubsection{Hyperparameters}
\begin{figure}[ht]
  \begin{center}
  \subfigure[]{\includegraphics[width=1.7in]{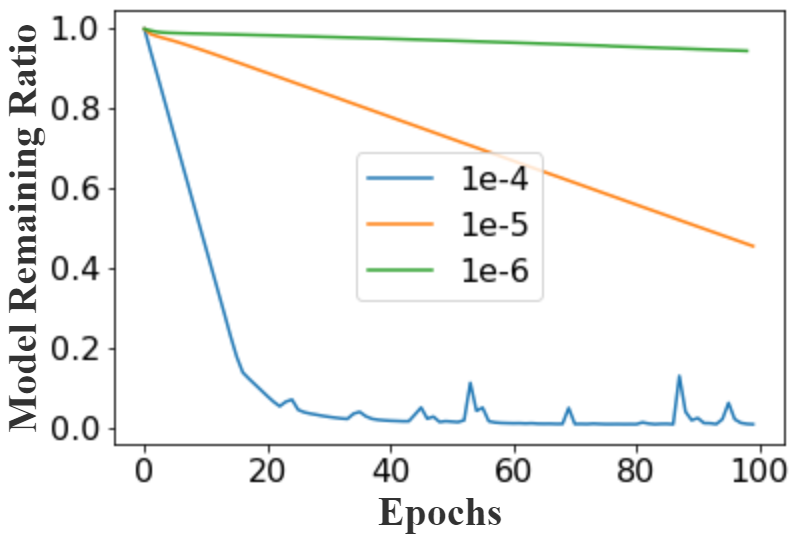}}
  \subfigure[]{\includegraphics[width=1.7in]{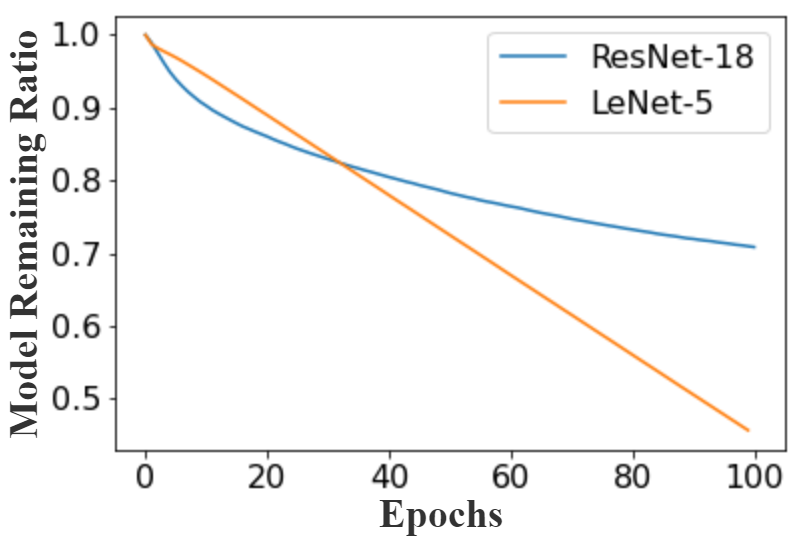}}
  \\
  \caption{(a) Pruning effect in ResNet-18 for different value of hyper-parameter $\alpha$. (b) Change of model remaining ratio with ResNet-18 and LeNet-5 for different $\alpha$. (Best viewed in color)}
 \label{figure10}
  \end{center}
\end{figure}

The hyperparameter $\alpha$ controls the intensity of pruning in the loss function during training, and it plays a crucial role in determining the final balance of model sparsity. We explore the effect of different $\alpha$ values, ranging from $10^{-3}$ to $10^{-7}$, in our experiments.
The value of $\alpha$ is mainly determined based on the data amount and training epochs, as pruning is primarily performed in each iteration of the training step. We aim to ensure that the product of the total number of iterations (e.g., image numbers multiplied by epochs) and $\alpha$ is approximately 1, allowing for efficient and effective pruning.
To investigate the influence of different $\alpha$ values on the model's pruning intensity, we conduct experiments using different models, such as ResNet-18 and LeNet-5. Additionally, we examine the change in the model's remaining ratio for various $\alpha$ values. The results are shown in Figure \ref{figure10}, providing insights into the relationship between $\alpha$, pruning intensity, and the model's remaining parameters for each architecture.

\section{Conclusion}

In this study, we aimed to tackle the challenge of managing heterogeneous datasets in continual learning. We observed unstable performance of rehearsal, regularization, and non-adaptive parameter isolation-based methods when dealing with multiple heterogeneous datasets in experiments.
Inspired by the neural-reuse principle of human brains, we presented AdaptCL, a novel continual learning algorithm. Our proposed method effectively addresses the challenge of managing heterogeneous datasets in continual learning, outperforming existing approaches in terms of robustness and achieving higher average accuracy. Additionally, AdaptCL proves to be a proficient few-shot learner, exhibiting the capability to make generalizations based on limited examples similar to human cognitive abilities. By introducing fine-grained data-driven pruning and task-agnostic parameter isolation, we address catastrophic forgetting and demonstrate the effectiveness of AdaptCL across heterogeneous datasets in diverse applications. Our work contributes to the field by providing a novel algorithm that improves performance in heterogeneous dataset scenarios. 
While our approach is computationally efficient, we acknowledge the limitation of reduced learning efficiency with insufficient model capacity. To address this, future work will focus on introducing network expansion techniques to enhance scalability on a growing number of heterogeneous datasets.

\section*{Acknowledgment}
This work was supported by Shenzhen-Hong Kong-Macau Technology Research Programme No. SGDX20201103095203029, HK RGC Research Impact Fund No. R5034-18, HK RGC General Research Fund No. PolyU 15204921, and Research Institute for Artificial Intelligence of Things, The Hong Kong Polytechnic University.

\ifCLASSOPTIONcaptionsoff
  \newpage
\fi



\bibliographystyle{IEEEtran}
\bibliography{IEEEabrv,main}

\begin{thebibliography}{10}
\providecommand{\url}[1]{#1}
\csname url@samestyle\endcsname
\providecommand{\newblock}{\relax}
\providecommand{\bibinfo}[2]{#2}
\providecommand{\BIBentrySTDinterwordspacing}{\spaceskip=0pt\relax}
\providecommand{\BIBentryALTinterwordstretchfactor}{4}
\providecommand{\BIBentryALTinterwordspacing}{\spaceskip=\fontdimen2\font plus
\BIBentryALTinterwordstretchfactor\fontdimen3\font minus \fontdimen4\font\relax}
\providecommand{\BIBforeignlanguage}[2]{{%
\expandafter\ifx\csname l@#1\endcsname\relax
\typeout{** WARNING: IEEEtran.bst: No hyphenation pattern has been}%
\typeout{** loaded for the language `#1'. Using the pattern for}%
\typeout{** the default language instead.}%
\else
\language=\csname l@#1\endcsname
\fi
#2}}
\providecommand{\BIBdecl}{\relax}
\BIBdecl
\renewcommand{\BIBentryALTinterwordstretchfactor}{4}

\bibitem{he2021online}
J.~He and F.~Zhu, ``Online continual learning for visual food classification,'' in \emph{Proceedings of the IEEE/CVF international conference on computer vision}, 2021, pp. 2337--2346.

\bibitem{zhao2021memory}
H.~Zhao, H.~Wang, Y.~Fu, F.~Wu, and X.~Li, ``Memory-efficient class-incremental learning for image classification,'' \emph{IEEE Transactions on Neural Networks and Learning Systems}, vol.~33, no.~10, pp. 5966--5977, 2021.

\bibitem{he2022exemplar}
J.~He and F.~Zhu, ``Exemplar-free online continual learning,'' in \emph{2022 IEEE International Conference on Image Processing (ICIP)}.\hskip 1em plus 0.5em minus 0.4em\relax IEEE, 2022, pp. 541--545.

\bibitem{abati2020conditional}
D.~Abati, J.~Tomczak, T.~Blankevoort, S.~Calderara, R.~Cucchiara, and B.~E. Bejnordi, ``Conditional channel gated networks for task-aware continual learning,'' in \emph{Proceedings of the IEEE/CVF Conference on Computer Vision and Pattern Recognition}, 2020, pp. 3931--3940.

\bibitem{zhao2022memory}
Y.~Zhao, D.~Saxena, and J.~Cao, ``Memory-efficient domain incremental learning for internet of things,'' in \emph{Proceedings of the 20th ACM Conference on Embedded Networked Sensor Systems}, 2022, pp. 1175--1181.

\bibitem{pascual2005plastic}
A.~Pascual-Leone, A.~Amedi, F.~Fregni, and L.~B. Merabet, ``The plastic human brain cortex,'' \emph{Annu. Rev. Neurosci.}, vol.~28, pp. 377--401, 2005.

\bibitem{johnston2009plasticity}
M.~V. Johnston, ``Plasticity in the developing brain: implications for rehabilitation,'' \emph{Developmental disabilities research reviews}, vol.~15, no.~2, pp. 94--101, 2009.

\bibitem{anderson2010neural}
M.~L. Anderson, ``Neural reuse: A fundamental organizational principle of the brain,'' \emph{Behavioral and brain sciences}, vol.~33, no.~4, pp. 245--266, 2010.

\bibitem{lopez2017gradient}
D.~Lopez-Paz and M.~Ranzato, ``Gradient episodic memory for continual learning,'' \emph{Advances in neural information processing systems}, vol.~30, 2017.

\bibitem{chaudhry2018efficient}
A.~Chaudhry, M.~Ranzato, M.~Rohrbach, and M.~Elhoseiny, ``Efficient lifelong learning with a-gem,'' \emph{arXiv preprint arXiv:1812.00420}, 2018.

\bibitem{10197260}
J.~Peng, D.~Ye, B.~Tang, Y.~Lei, Y.~Liu, and H.~Li, ``Lifelong learning with cycle memory networks,'' \emph{IEEE Transactions on Neural Networks and Learning Systems}, pp. 1--14, 2023.

\bibitem{10058177}
S.~Ho, M.~Liu, L.~Du, L.~Gao, and Y.~Xiang, ``Prototype-guided memory replay for continual learning,'' \emph{IEEE Transactions on Neural Networks and Learning Systems}, pp. 1--11, 2023.

\bibitem{li2017learning}
Z.~Li and D.~Hoiem, ``Learning without forgetting,'' \emph{IEEE transactions on pattern analysis and machine intelligence}, vol.~40, no.~12, pp. 2935--2947, 2017.

\bibitem{rosasco2022distilled}
A.~Rosasco, A.~Carta, A.~Cossu, V.~Lomonaco, and D.~Bacciu, ``Distilled replay: Overcoming forgetting through synthetic samples,'' in \emph{Continual Semi-Supervised Learning: First International Workshop, CSSL 2021, Virtual Event, August 19--20, 2021, Revised Selected Papers}.\hskip 1em plus 0.5em minus 0.4em\relax Springer, 2022, pp. 104--117.

\bibitem{sun2020distill}
J.~Sun, S.~Wang, J.~Zhang, and C.~Zong, ``Distill and replay for continual language learning,'' in \emph{Proceedings of the 28th international conference on computational linguistics}, 2020, pp. 3569--3579.

\bibitem{rebuffi2017icarl}
S.-A. Rebuffi, A.~Kolesnikov, G.~Sperl, and C.~H. Lampert, ``icarl: Incremental classifier and representation learning,'' in \emph{Proceedings of the IEEE conference on Computer Vision and Pattern Recognition}, 2017, pp. 2001--2010.

\bibitem{binici2022robust}
K.~Binici, S.~Aggarwal, N.~T. Pham, K.~Leman, and T.~Mitra, ``Robust and resource-efficient data-free knowledge distillation by generative pseudo replay,'' in \emph{Proceedings of the AAAI Conference on Artificial Intelligence}, vol.~36, no.~6, 2022, pp. 6089--6096.

\bibitem{kirkpatrick2017overcoming}
J.~Kirkpatrick, R.~Pascanu, N.~Rabinowitz, J.~Veness, G.~Desjardins, A.~A. Rusu, K.~Milan, J.~Quan, T.~Ramalho, A.~Grabska-Barwinska \emph{et~al.}, ``Overcoming catastrophic forgetting in neural networks,'' \emph{Proceedings of the national academy of sciences}, vol. 114, no.~13, pp. 3521--3526, 2017.

\bibitem{huszar2017quadratic}
F.~Husz{\'a}r, ``On quadratic penalties in elastic weight consolidation,'' \emph{arXiv preprint arXiv:1712.03847}, 2017.

\bibitem{9705128}
Y.~Liu, X.~Hong, X.~Tao, S.~Dong, J.~Shi, and Y.~Gong, ``Model behavior preserving for class-incremental learning,'' \emph{IEEE Transactions on Neural Networks and Learning Systems}, vol.~34, no.~10, pp. 7529--7540, 2023.

\bibitem{schwarz2018progress}
J.~Schwarz, W.~Czarnecki, J.~Luketina, A.~Grabska-Barwinska, Y.~W. Teh, R.~Pascanu, and R.~Hadsell, ``Progress \& compress: A scalable framework for continual learning,'' in \emph{International Conference on Machine Learning}.\hskip 1em plus 0.5em minus 0.4em\relax PMLR, 2018, pp. 4528--4537.

\bibitem{li2020continual}
H.~Li, P.~Barnaghi, S.~Enshaeifar, and F.~Ganz, ``Continual learning using bayesian neural networks,'' \emph{IEEE transactions on neural networks and learning systems}, vol.~32, no.~9, pp. 4243--4252, 2020.

\bibitem{park2020convolutional}
G.-M. Park, S.-M. Yoo, and J.-H. Kim, ``Convolutional neural network with developmental memory for continual learning,'' \emph{IEEE Transactions on Neural Networks and Learning Systems}, vol.~32, no.~6, pp. 2691--2705, 2020.

\bibitem{serra2018overcoming}
J.~Serra, D.~Suris, M.~Miron, and A.~Karatzoglou, ``Overcoming catastrophic forgetting with hard attention to the task,'' in \emph{International Conference on Machine Learning}.\hskip 1em plus 0.5em minus 0.4em\relax PMLR, 2018, pp. 4548--4557.

\bibitem{ororbia2020continual}
A.~Ororbia, A.~Mali, C.~L. Giles, and D.~Kifer, ``Continual learning of recurrent neural networks by locally aligning distributed representations,'' \emph{IEEE Transactions on Neural Networks and Learning Systems}, vol.~31, no.~10, pp. 4267--4278, 2020.

\bibitem{rusu2016progressive}
A.~A. Rusu, N.~C. Rabinowitz, G.~Desjardins, H.~Soyer, J.~Kirkpatrick, K.~Kavukcuoglu, R.~Pascanu, and R.~Hadsell, ``Progressive neural networks,'' \emph{arXiv preprint arXiv:1606.04671}, 2016.

\bibitem{10036133}
R.~Ma, Q.~Wu, K.~N. Ngan, H.~Li, F.~Meng, and L.~Xu, ``Forgetting to remember: A scalable incremental learning framework for cross-task blind image quality assessment,'' \emph{IEEE Transactions on Multimedia}, pp. 1--12, 2023.

\bibitem{xu2018reinforced}
J.~Xu and Z.~Zhu, ``Reinforced continual learning,'' \emph{Advances in Neural Information Processing Systems}, vol.~31, 2018.

\bibitem{adel2019continual}
T.~Adel, H.~Zhao, and R.~E. Turner, ``Continual learning with adaptive weights (claw),'' \emph{arXiv preprint arXiv:1911.09514}, 2019.

\bibitem{fernando2017pathnet}
C.~Fernando, D.~Banarse, C.~Blundell, Y.~Zwols, D.~Ha, A.~A. Rusu, A.~Pritzel, and D.~Wierstra, ``Pathnet: Evolution channels gradient descent in super neural networks,'' \emph{arXiv preprint arXiv:1701.08734}, 2017.

\bibitem{rajasegaran2019random}
J.~Rajasegaran, M.~Hayat, S.~Khan, F.~S. Khan, and L.~Shao, ``Random path selection for incremental learning,'' \emph{Advances in Neural Information Processing Systems}, 2019.

\bibitem{ke2020continual}
Z.~Ke, B.~Liu, and X.~Huang, ``Continual learning of a mixed sequence of similar and dissimilar tasks,'' \emph{Advances in Neural Information Processing Systems}, vol.~33, pp. 18\,493--18\,504, 2020.

\bibitem{rosenfeld2018incremental}
A.~Rosenfeld and J.~K. Tsotsos, ``Incremental learning through deep adaptation,'' \emph{IEEE transactions on pattern analysis and machine intelligence}, vol.~42, no.~3, pp. 651--663, 2018.

\bibitem{golkar2019continual}
S.~Golkar, M.~Kagan, and K.~Cho, ``Continual learning via neural pruning,'' \emph{arXiv preprint arXiv:1903.04476}, 2019.

\bibitem{mallya2018packnet}
A.~Mallya and S.~Lazebnik, ``Packnet: Adding multiple tasks to a single network by iterative pruning,'' in \emph{Proceedings of the IEEE conference on Computer Vision and Pattern Recognition}, 2018, pp. 7765--7773.

\bibitem{delange2021continual}
M.~Delange, R.~Aljundi, M.~Masana, S.~Parisot, X.~Jia, A.~Leonardis, G.~Slabaugh, and T.~Tuytelaars, ``A continual learning survey: Defying forgetting in classification tasks,'' \emph{IEEE Transactions on Pattern Analysis and Machine Intelligence}, 2021.

\bibitem{aljundi2018memory}
R.~Aljundi, F.~Babiloni, M.~Elhoseiny, M.~Rohrbach, and T.~Tuytelaars, ``Memory aware synapses: Learning what (not) to forget,'' in \emph{Proceedings of the European Conference on Computer Vision (ECCV)}, 2018, pp. 139--154.

\bibitem{liu2020dynamic}
J.~Liu, Z.~Xu, R.~Shi, R.~C. Cheung, and H.~K. So, ``Dynamic sparse training: Find efficient sparse network from scratch with trainable masked layers,'' \emph{arXiv preprint arXiv:2005.06870}, 2020.

\bibitem{xu2019accurate}
Z.~Xu and R.~C. Cheung, ``Accurate and compact convolutional neural networks with trained binarization,'' \emph{arXiv preprint arXiv:1909.11366}, 2019.

\bibitem{peng2019moment}
X.~Peng, Q.~Bai, X.~Xia, Z.~Huang, K.~Saenko, and B.~Wang, ``Moment matching for multi-source domain adaptation,'' in \emph{Proceedings of the IEEE International Conference on Computer Vision}, 2019, pp. 1406--1415.

\bibitem{bottou1991stochastic}
L.~Bottou \emph{et~al.}, ``Stochastic gradient learning in neural networks,'' \emph{Proceedings of Neuro-N{\i}mes}, vol.~91, no.~8, p.~12, 1991.

\end{thebibliography}

\vfill


\end{document}